\documentclass[acmtog, authorversion, nonacm]{acmart}

\AtBeginDocument{%
  \providecommand\BibTeX{{%
    \normalfont B\kern-0.5em{\scshape i\kern-0.25em b}\kern-0.8em\TeX}}}

\acmJournal{FACMP}

\usepackage{graphicx}
\usepackage{tikz}
\usepackage{comment}

\usepackage{color}

\usepackage{array}
\usepackage{tabu}
\usepackage{array}
\usepackage{booktabs}
\usepackage{soul}

\usepackage{xcolor}

\usepackage{bbm}

\newif\ifdraft
\draftfalse

\ifdraft
\newcommand{\okc}[1]{{\color{orange}[\textbf{OK:} #1]}}
\newcommand{\dcc}[1]{{\color{purple}[\textbf{DC:} #1]}}
\newcommand{\dlc}[1]{{\color{blue}[\textbf{DL:} #1]}}

\else
\newcommand{\okc}[1]{}
\newcommand{\dcc}[1]{}
\newcommand{\dlc}[1]{}

\fi

\definecolor{mygray}{RGB}{140, 140, 140}

\definecolor{dashedpurple}{RGB}{118,0,103}
\definecolor{dashedorange}{RGB}{181,100,13}

\newcommand{\Loss}{\mathcal{L}}
\newcommand{\Rspace}{\mathbb{R}}

\newcommand{\norm}[1]{\left\lVert#1\right\rVert}
\newcommand{\innerprod}[2]{{\left\langle#1,#2\right\rangle}}
\DeclareMathOperator*{\argmax}{arg\,max}

\makeatletter
\newcommand{\thickhline}{%
    \noalign {\ifnum 0=`}\fi \hrule height 1pt
    \futurelet \reserved@a \@xhline
}
\makeatletter
\newcommand{\thickvline}{%
    \noalign {\ifnum 0=`}\fi \vrule height 1pt
    \futurelet \reserved@a \@xvline
}
\newcolumntype{"}{@{\hskip\tabcolsep\vrule width 1pt\hskip\tabcolsep}}
\makeatother

\newcommand\figref[1]{Fig.~\ref{#1}}

\DeclareMathAlphabet\mathbfcal{OMS}{cmsy}{b}{n}

\newcommand{\SE}{$\mathbf{SE}(3)$}
\newcommand{\SO}{$\mathbf{SO}(3)$}

\setcitestyle{nosort,square}
\begin{document}

\title{Shape-Pose Disentanglement using \SE-equivariant Vector Neurons}

\author{Oren Katzir}
\affiliation{\institution{Tel Aviv University} }
\author{Dani Lischinski}
\affiliation{\institution{Hebrew University of Jerusalem}} 
\author{Daniel Cohen-Or}
\affiliation{\institution{Tel Aviv University} }

\begin{abstract}
We introduce an unsupervised technique for encoding point clouds into a canonical shape representation, by disentangling shape and pose.
Our encoder is stable and consistent, meaning that the shape encoding is purely pose-invariant, while the extracted rotation and translation are able to semantically align different input shapes of the same class to a common canonical pose.
Specifically, we design an auto-encoder based on Vector Neuron Networks, a rotation-equivariant neural network, whose layers we extend to provide  translation-equivariance in addition to rotation-equivariance only.
The resulting encoder produces pose-invariant shape encoding by construction, enabling our approach to focus on learning a consistent canonical pose for a class of objects.
Quantitative and qualitative experiments validate the superior stability and consistency of our approach.

\end{abstract}

%\begin{teaserfigure}
  %\includegraphics[width=\textwidth]{figures/teaser.pdf}
  %\caption{caption.}
  %\label{fig:teaser}
%\end{teaserfigure}

\maketitle

\newcommand\offsetx{.93}
\newcommand\offsety{.39}

\section{Introduction}
\label{sec:Intro}

Point clouds reside at the very core of 3D geometry processing, as they are acquired at the beginning of the 3D processing pipeline and usually serve as the raw input for shape analysis or surface reconstruction. Thus, understanding the underlying geometry of a point cloud has a profound impact on the entire 3D processing chain. This task, however, is challenging since point clouds are unordered, and contain neither connectivity, nor any other global information.

In recent years, with the emergence of neural networks, various techniques have been developed to circumvent the challenges of analyzing and understanding point clouds~\cite{qi2017pointnet,qi2017pointnet++,wang2019dynamic,hamdi2021mvtn,ma2022rethinking,shi2019pointrcnn,liu2020tanet,yang20203dssd}. 
However, most methods rely on \emph{pre-aligned} datasets, where the point clouds are normalized, translated and oriented to have the same pose. 

In this work, we present an unsupervised technique to learn a canonical shape representation by disentangling shape, translation, and rotation.
Essentially, the canonical representation is required to meet two conditions: \emph{stability} and \emph{consistency}. The former means that the shape encoding should be invariant to any rigid transformation of the same input, while the latter means that different shapes of the same class should be semantically aligned, sharing the same canonical pose.
%First, the canonical representation should be \emph{stable}, i.e., it should be the same for the same input shape, regardless of pose. Second, it should be {\it consistent} across objects of the same class, i.e., they should be aligned.

Canonical alignment is not a new concept. 
Recently, Canonical Capsules~\cite{sun2021canonical} and Compass~\cite{spezialetti2020learning} proposed self-supervised learning of canonical representations using augmentations with Siamese training. We discuss these methods in more detail in the next section.
%\ok{However, the underlying equivariant and invariant backbone in both is only so by augmentation.}
In contrast, our approach is to extract a \emph{pose-invariant} shape encoding, which is explicitly disentangled from the separately extracted translation and rotation. 
%Unlike these methods, our {\it shape} encoding is {\it pose} invariant by construction, explicitly disentangling shape from pose.

Specifically, we design an auto-encoder, trained on an unaligned dataset, that encodes the input point cloud into three disentangled components: (i) a pose-invariant shape encoding, (ii) a rotation  matrix and (iii) a translation vector. We achieve pure \SE-invariant shape encoding and \SE-equivariant pose estimation (enabling reconstruction of the input shape), by leveraging a novel extension of the recently proposed Vector Neuron Networks (VNN)~\cite{deng2021vector}. The latter is an \SO-equivariant neural network for point cloud processing, and while translation invariance could theoretically be achieved by centering the input point clouds, such approach is sensitive to noise, missing data and partial shapes. Therefore we propose an extension to VNN  achieving \SE-equivariance. 
% whose components we extend to achieve \SE-equivariance.

It should be noted that the shape encodings produced by our network are stable (i.e., pose-invariant) \emph{by construction}, due to the use of \SE-invariant layers.

At the same time, the extracted rigid transformation is equivariant to the pose of the input. This enables the learning process to focus on the consistency across different shapes. Consistency is achieved by altering the input point cloud with a variety of simple shape augmentations, while keeping the pose fixed, allowing us to constrain the learned transformation to be invariant to the identity, (i.e., the particular shape), of the input point cloud.

Moreover, our disentangled shape and pose representation is not limited to point cloud decoding, but can be combined with any 3D data decoder, as we demonstrate by learning a canonical implicit representation of our point cloud utilizing occupancy networks~\cite{mescheder2019occupancy}.

We show, both qualitatively and quantitatively, that our approach leads to a stable, consistent, and purely \SE-invariant canonical representation compared to previous approaches.

 %Moreover, as we show, since canonical capsules is only invariant by augmentation it exhibits noisy results for out-of-distribution point clouds, while compass relies on a semi-equivariant layers, which also results unstable canonical framework which is input transformation depended. 

\section{Background and Related Work}
\subsection{Canonical representation}
A number of works proposed techniques to achieve learnable canonical frames, typically requiring some sort of supervision \cite{rempe2020caspr,novotny2019c3dpo,gu2020weakly}.
%C3DPO~\cite{novotny2019c3dpo} used multiple views of moving keypoints (i.e. motion of joints) to supervise a factorization of structure and motion.
Recently, two unsupervised methods were proposed: Canonical Capsules~\cite{sun2021canonical} and Compass~\cite{spezialetti2020learning}. 
Canonical Capsules~\cite{sun2021canonical} is an auto-encoder network that extracts positions and pose-invariant descriptors for $k$ capsules, from which the input shape may be reconstructed. Pose invariance and equivariance are achieved only implicitly via Siamese training, by feeding the network with pairs of rotated and translated versions of the same input point cloud.
%However, the underlying encoder~\cite{sun2020acne} is only invariant by augmentation.

Compass~\cite{spezialetti2020learning} builds upon spherical CNN~\cite{cohen2018spherical}, a semi-equivariant \SO{} network, to estimate the pose with respect to the canonical representation.
It should be noted that Compass is inherently tied to spherical CNN, which is not purely equivariant~\cite{cohen2018spherical}. Thus, similarly to Canonical Capsules, Compass augments the input point cloud with a rotated version to regularize an equivariant pose estimation. It should be noted that neither method guarantees pure equivariance. 

Similarly to Canonical Capsules, we employ an auto-encoding scheme to disentangle pose from shape, i.e., the canonical representation, and similarly to Compass, we strive to employ an equivariant network, however, our network is \SE-equivariant and not only \SO-equivariant.
More importantly, differently from these two approaches, the different branches of our network are \SE-invariant or \SE-equivariant \emph{by construction}, and thus the learning process is free from the burden of enforcing these properties. Rather, the process focuses on learning a consistent shape representation in a canonical pose.
%learning has no relation whatsoever to the network being invariant or equivariant, since our method is by construction both equivariant and invariant to SE(3) transformations, regardless of the training process. Rather, we explicitly optimize for a consistent canonical representation.

\subsection{3D reconstruction}
Our method reconstructs an input point cloud by disentangling the input 3D geometry into shape and pose. The encoder outputs a pose encoding and a shape encoding which is pose-invariant by construction, while the decoder reconstructs the 3D geometry from the shape encoding alone. Consequently, our architecture can be easily integrated into various 3D auto-encoding pipelines.
In this work, we shall demonstrate our shape-pose disentanglement for point cloud encoding and implicit representation learning.

State-of-the-art point cloud auto-encoding methods rely on a folding operation of a template (optionally learned) hyperspace point cloud  to the input 3D point cloud~\cite{yang2018foldingnet,groueix2018papier,deprelle2019learning}.  Following this approach, we employ AtlasNetV2~\cite{deprelle2019learning} which uses multiple folding operations from hyperspace patches to 3D coordinates, to reconstruct point clouds in a pose-invariant frame.

Implicit 3D representation networks~\cite{mescheder2019occupancy,park2019deepsdf,xu2019disn} enable learning of the input geometry with high resolution and different mesh topology.  We utilize occupancy networks~\cite{mescheder2019occupancy} to learn an implicit pose-invariant shape representation.

\subsection{ Rotation-equivariance and Vector Neuron Network}
\label{sec:VNN}
The success of 2D convolutional neural networks (CNN) on images, which are equivariant to translation, drove a similar approach for 3D data with rotation as the symmetry group.
The majority of works on 3D rotation-equivariance~\cite{esteves2018learning,cohen2018spherical,thomas2018tensor,weiler20183d}, focus on steerable CNNs~\cite{cohen2016steerable}, where each layer ``steers'' the output features according to the symmetry property (rotation and occasionally translation for 3D data).
For example, Spherical CNNs~\cite{esteves2018learning,cohen2018spherical} transform the input point cloud to a spherical signal, and use spherical harmonics filters, yielding features on \SO-space.
%TensorField~\cite{} also uses spherical harmonics filters, however the signal is not limited to spherical one, as each layer's input a scalar, a vector, and a higher-order tensors. 
Usually, these methods are tied with specific architecture design and data input which limit their applicability and adaptation to SOTA 3D processing. 

Recently, Deng et al.~\cite{deng2021vector} introduced Vector Neuron Networks (VNN), a rather light and elegant framework for \SO-equivariance. Empirically, the VNN design performs on par with more complex and specific architectures.
The key benefit of VNNs lies in their simplicity, accessibility and generalizability. Conceptually, any standard point cloud processing network can be elevated to \SO-equivariance (and invariance) with minimal changes to its architecture.
Below we briefly describe VNNs and refer the reader to~\cite{deng2021vector} for further details.

In VNNs the representation of a single neuron is lifted from a sequence of scalar values to a sequence of 3D vectors.
%We denote a point cloud with $N$ points by $\bold{X} \in \Rspace^{N \times 3}$ and the vector neuron features by $\mathcal{V} \in \Rspace^{N \times C \times 3}$, where $C$ is the number of channels. A single point feature is then a matrix $\bold{V} \in \Rspace^{C \times 3}$. 
A single vector neuron feature is thus a matrix $\bold{V} \in \Rspace^{C \times 3}$, and we denote a collection of $N$ such features by $\mathcal{V} \in \Rspace^{N \times C \times 3}$.
The layers of VNNs, which map between such collections, $f: \mathcal{V} \in \Rspace^{N \times C \times 3} \rightarrow \bold{\mathcal{V}'} \in \Rspace^{N \times C' \times 3}$, are equivariant to rotations $R \in \Rspace^{3 \times 3}$, that is:
\begin{equation}
	f\left(\mathcal{V} R\right) = f\left(\bold{\mathcal{V}}\right)R,
\end{equation}
where $\mathcal{V} R = \left\{\bold{V}_n R\right\}_{n=1}^N$.

Ordinary linear layers fulfill this requirement, however, other non-linear layers, such as ReLU and max-pooling, do not.
For ReLU activation, VNNs apply a truncation w.r.t to a learned half-space. Let $\bold{V}, \bold{V'} \in  \Rspace^{C \times 3}$ be the input and output vector neuron features of a single point, respectively. Each 3D vector $\bold{v}' \in \bold{V}'$ is obtained by first applying two learned matrices $\bold{Q}, \bold{K} \in \Rspace^{1 \times C}$ to project $\bold{V}$ to a feature $\bold{q} = \bold{Q}\bold{V} \in \Rspace^{1 \times 3}$ and a direction $\bold{k}=\bold{K}\bold{V} \in \Rspace^{1 \times 3}$.
To achieve equivariance, $\bold{v}' \in\bold{ V'}$ is then defined by truncating the part of $\bold{q}$ that lies in the negative half-space of $\bold{k}$, as follows,
\begin{equation}
  \bold{v'} =\begin{cases}
    \bold{q} & \text{if} \quad \innerprod{\bold{q}}{\bold{k}} \geq 0, \\
    \bold{q} - \innerprod{\bold{q}}{\frac{\bold{k}}{\norm{\bold{k}}}}\frac{\bold{k}}{\norm{\bold{k}}} & \text{otherwise}.
  \end{cases}
\end{equation}
In addition, VNNs employ rotation-equivariant pooling operations and normalization layers. We refer the reader to \cite{deng2021vector} for the complete definition.

Invariance layers can be achieved by inner product of two rotation-equivariant features. Let $\bold{V} \in \Rspace^{C \times 3}$, and  $\bold{V'} \in \Rspace^{C' \times 3}$ be two equivariant features obtained from an input point cloud $\bold{X}$.
Then rotating $\bold{X}$ by a matrix $R$, results in the features $\bold{V}R$ and $\bold{V'}R$, and
\begin{equation}
	\langle \bold{V}R,\bold{V'}R\rangle = \bold{V}R(\bold{V'}R)^T = \bold{V}RR^T\bold{V'}^T = \bold{V}\bold{V'}^T =   \langle \bold{V},\bold{V'}\rangle.
\end{equation}

In our work, we also utilize vector neurons, but we extend the different layers to be \SE-equivariant, instead of \SO-equivariant, as described in Section~\ref{subsec:VNT}. This new design allow us to construct an \SE-invariant encoder, which gradually disentangles the pose from the shape, first the translation and then the rotation, resulting in a pose-invariant shape encoding.

\begin{figure}
	\centering
	\includegraphics[width=1.0\linewidth]{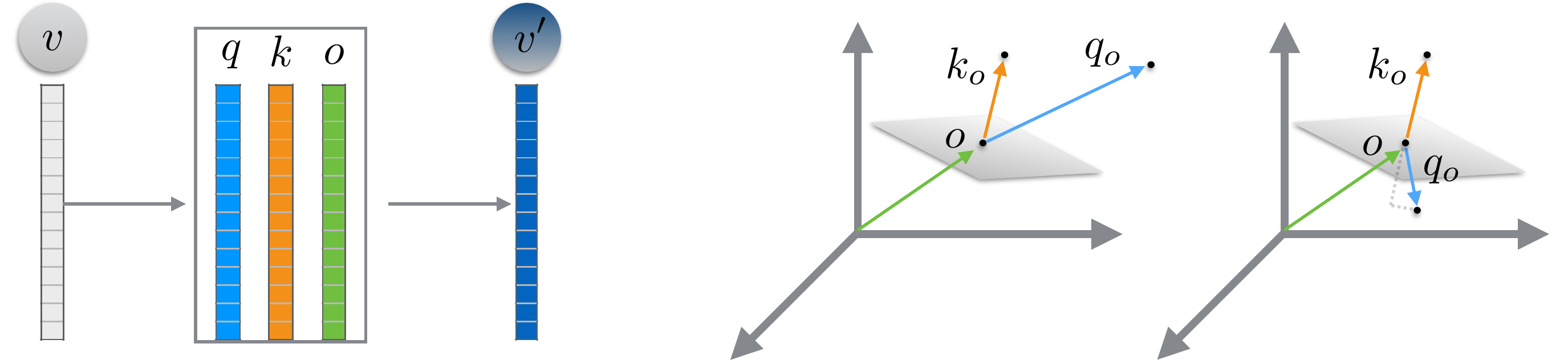}
	\caption{Vector Neuron Translation equivariant non linear layer.
	We learn for each input point feature $\bold{v}$, three component $\bold{o},\bold{q}, \bold{k}$, and interpret them as an origin $\bold{o}$, a feature $\bold{q_o} = \bold{q} - \bold{o}$ and a direction $\bold{k_o} = \bold{k} - \bold{o}$. Similarly to VNN operation, the feature component of $\bold{q_o}$ which is in the half-space defined by $-\bold{k_o}$ is clipped. In-addition, we translate the feature by the learned origin $\bold{o}$ outputting the $\bold{v'}$.}
	\label{fig:vnt}
\end{figure}

\section{Method}
\label{sec:method}

We design an auto-encoder to disentangle shape, translation, and rotation. We wish the resulting representation to be stable, i.e., the shape encoding should be pose-invariant, and the pose \SE-equivariant.
At the same we wish multiple different shapes in the same class to have a consistent canonical pose.
To achieve stability, we revisit VNNs and design new \SE-equivariant and invariant layers, which we refer to as Vector Neurons with Translation (VNT). Consistency is then achieved by self-supervision, designed to preserve pose across shapes.
In the following, we first describe the design of our new VNT layers. Next, we present our VNN and VNT-based auto-encoder architecture. Finally, we elaborate on our losses to encourage disentanglement of shape from pose in a consistent manner.

\subsection{\SE-equivariant Vector Neuron Network}
\label{subsec:VNT}

As explained earlier, Vector Neuron Networks (VNN) \cite{deng2021vector} provide a framework for \SO-equivariant and invariant point cloud processing.
Since a pose of an object consists of translation and rotation, 
\SE-equivariance and invariance are needed for shape-pose disentanglement. 
While it might seem that centering the input point cloud should suffice, note that point clouds are often captured with noise and occlusions, leading to missing data and partial shapes, which may significantly affect the global center of the input.
Specifically, for canonical representation learning, a key condition is consistency across different objects, thus, such an approach assumes that the center of the point cloud is consistently semantic between similar but different objects, which is hardly the case.
Equivariance to translation, on the other-hand, allows identifying local features in different locations with the same filters, without requiring global parameters.

Therefore, we revisit the Vector Neuron layers and extend them to Vector Neurons with Translation (VNT), thereby achieving \SE-equivariance.

\subsubsection{Linear layers:}
While linear layers are by definition rotation-equivariant, they are not translation-equivariant.
Following VNN, our linear module $f_{\textrm{lin}}\left(\cdot;\bold{W}\right)$ is defined via a weight matrix $\bold{W} \in \Rspace^{C' \times C}$, acting on a vector-list feature $\bold{V} \in \Rspace^{C \times 3}$. Let $R \in \Rspace^{3 \times 3}$ be a rotation matrix and $T \in \Rspace^{1\times 3}$ a translation vector.
For  $f_{\textrm{lin}}\left(\cdot;\bold{W}\right)$  to be \SE-equivariant, the following must hold:
\begin{equation}
    f_{\textrm{lin}}\left(\bold{V}R + \mathbbm{1}_CT\right) = \bold{W}\bold{V}R + \bold{\mathbbm{1}}_{C'} T = f_{\textrm{lin}}(\bold{V})R + \bold{\mathbbm{1}}_{C'} T,
    \label{eq:lin_eq_T}
\end{equation}
where $\mathbbm{1}_C = [1,1,\dots, 1]^T \in \Rspace^{C \times 1}$ is a column vector of length $C$. 
A sufficient condition for \eqref{eq:lin_eq_T} to hold is achieved by constraining each row of $\bold{W}$ to sum to one. Formally, $\bold{W} \in \mathcal{W}^{C' \times C}$, where
\begin{equation}
\mathcal{W}^{C' \times C} = \left\{\bold{W} \in \Rspace^{C' \times C} \mid \textstyle{\sum_{j=1}^{C}} w_{i,j} = 1 \quad \forall i \in [1,C'] \right\},
\end{equation}
See the supplementary material for a complete proof.

\subsubsection{Non-linear layers:}
We extend each non-linear VNN layer to become \SE-equivariant by adding a learnable origin.  
More formally, for the ReLU activation layer, given an input feature list $\bold{V} \in \Rspace^{C \times 3}$, we learn three (rather than two) linear maps,  $\bold{Q}, \bold{K}, \bold{O} \in \mathcal{W}^{1 \times C}$ projecting the input to $\bold{q}, \bold{k}, \bold{o} \in \Rspace^{1 \times 3}$.  The feature and direction are defined w.r.t the origin $\bold{o}$, i.e., the feature is given by $\bold{q_o} = \bold{q}-\bold{o}$, while the direction is given by $\bold{k_o} = \bold{k}-\bold{o}$, as illustrated in \figref{fig:vnt}. The ReLU is applied by clipping the part of $\bold{q_o}$ that resides behind the plane defined by $\bold{k_o}$ and $\bold{o}$, i.e.,
\begin{equation}
  \bold{v'} =\begin{cases}
    \bold{o} + \bold{q_o} & \text{if} \quad \innerprod{\bold{q_o}}{\bold{k_o}} \geq 0, \\
    \bold{o} + \bold{q_o} - \innerprod{\bold{q_o}}{\frac{\bold{k_o}}{\norm{\bold{k_o}}}}\frac{\bold{k_o}}{\norm{\bold{k_o}}} , & \text{otherwise}.
  \end{cases},
\end{equation}
Note that $\bold{o} + \bold{q_o} = \bold{q}$, and that $ \bold{K}, \bold{O}$ may be shared across the elements of $\bold{V}.$

%\okc{ Previously was - 
%we learn three (rather than two) linear maps,  $\bold{Q}, \bold{K}, \bold{O} \in \mathcal{W}^{1 \times C}$. 
%Similarly to VNN, these linear maps project the input to a feature $\bold{q}$, a direction $\bold{k}$, and an origin $\bold{o}$, such that the output of the extended activation layer is:

%\begin{equation}
%  \bold{v'} =\begin{cases}
%    \bold{q}  & \text{if} \quad \innerprod{\bold{q} - \bold{o}}{\bold{k} -\bold{o}} \geq 0, \\
%    \bold{q} - \innerprod{\bold{q}-\bold{o}}{\frac{\bold{k}-\bold{o}}{\norm{\bold{k}-\bold{o}}}}\frac{\bold{k}-\bold{o}}{\norm{\bold{k}-\bold{o}}} , & \text{otherwise}.
%  \end{cases}
%\end{equation},
%}

It may be easily seen that we preserve the equivariance w.r.t \SO{} rotations, as well translations. In the same manner, we extend the \SO-equivariant VNN maxpool layer to become \SE-equivariant.
We refer the reader to the supplementary material for the exact adaptation and complete proof. 

\subsubsection{Translation-invariant layers:}
\label{subsubsec:invarinat_trans_layer}
Invariance to translation can be achieved by subtracting two \SE-equivariant features.  Let $\bold{V},\bold{V'} \in \Rspace^{C \times 3}$ be two \SE-equivariant features obtained from an input point cloud $\bold{X}$. Then, rotating $\bold{X}$ by a matrix $R$ and translating by $T$, results in the features $\bold{V}R + \bold{\mathbbm{1}}_{C}T$ and $\bold{V'}R + \bold{\mathbbm{1}}_{C}T$, whose difference is translation-invariant:
\begin{equation}
	\left(\bold{V}R + \bold{\mathbbm{1}}_CT\right) - \left(\bold{V'}R + \bold{\mathbbm{1}}_CT\right) = \left(\bold{V} - \bold{V'}\right)R
\end{equation}
Note that the resulting feature is still rotation-equivariant, which enables to process it with VNN layers, further preserving \SO-equivariance.

%Please note that an equivariant SE(3) features $V'$ can be achieved by mean-pool operation on $V$ across the points index, i.e., $V' = \frac{1}{N}\sum_i{\bold{V}_i}$

\subsection{\SE-equivariant Encoder-Decoder}
We design an auto-encoder based on VNT and VNN layers to disentangle pose from shape. Thus, our shape representation is pose-invariant (i.e., stable), while our pose estimation is \SE-pose-equivariant, by construction. The decoder, which can be an arbitrary 3D decoder network, reconstructs the 3D shape from the invariant features.

\begin{figure}
	\centering
	\includegraphics[width=1.0\linewidth]{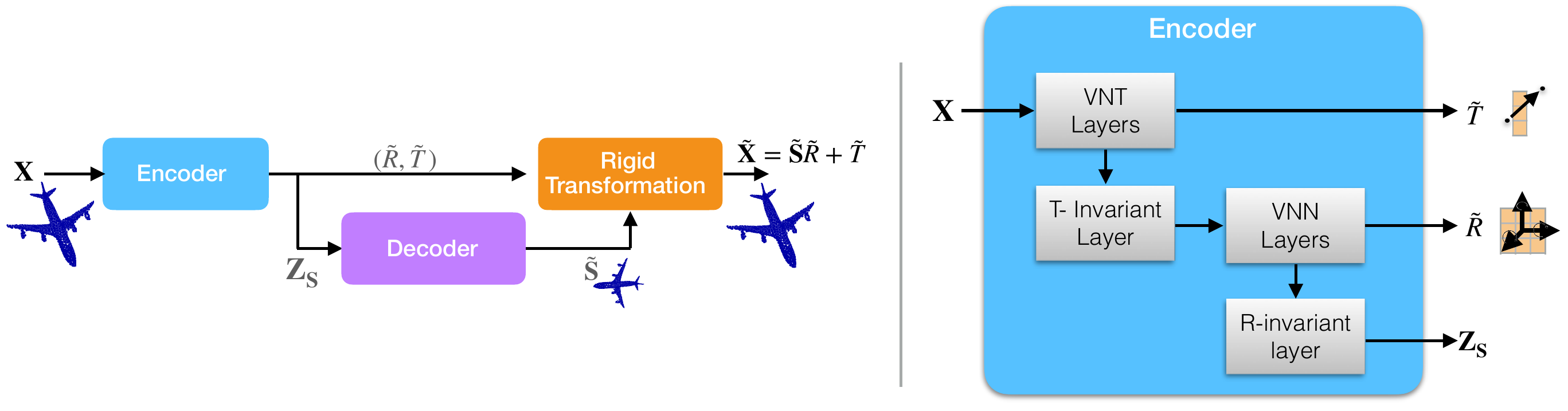}
	\caption{The architecture of our auto-encoder for shape-pose disentanglement. The auto-encoder (left) disentangles the input point cloud $\bold{X}$ to rotation $\tilde{R}$, translation $\tilde{T}$ and a canonical representation $\tilde{\bold{S}}$. The shape encoding $\bold{Z_s}$ is invariant by construction to the pose, while the learned rotation and translation are equivariant to it. Our encoder (right) learns features that are initially equivariant to the pose, and gradually become invariant to it, first to translation and then to rotation, eventually yielding pose invariant features $\bold{Z_s}$.}
	\label{fig:arch}
\end{figure}

\begin{figure*}[t]
	\centering
	\includegraphics[width=1.0\linewidth]{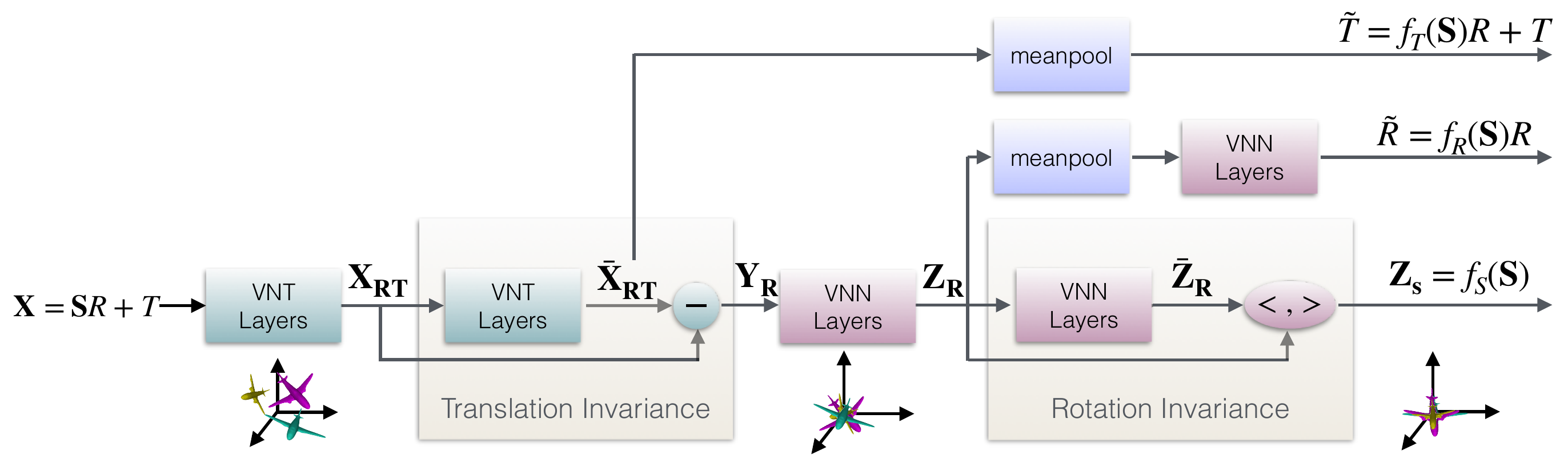}
	\caption{The architecture of our encoder. The representation of the input point cloud $\bold{X}$ yields a pose-invariant (bottom branch) shape encoding $\bold{Z_s}$ in two steps: first, making it invariant to translation and then to rotation. At the same time, the learned rigid transformation $(\tilde{R}, \tilde{T})$ (top and middle branches) is equivariant to the input pose. The small rendered planes at the bottom, illustrate the alignment at each stage.}
	\label{fig:enc_arch}
\end{figure*} 

The overall architecture of our AE is depicted in \figref{fig:arch}.
Given an input point cloud $\bold{X} \in \Rspace^{N \times 3}$, we can represent it as a rigid transformation of an unknown canonical representation $\bold{S} \in \Rspace^{N \times 3}$:
\begin{equation}
	\bold{X} = \bold{S}R + \mathbbm{1}_NT,
\end{equation}
where $\mathbbm{1}_N = [1,1,\dots, 1]^T \in \Rspace^{N \times 1}$ is a column vector of length $N$, $R \in \Rspace^{3 \times 3}$ is a rotation matrix and $T\in \Rspace^{1 \times 3}$ is a translation vector.

Our goal is to find the shape $\bold{S}$, which is by definition pose-invariant and should be consistently aligned across different input shapes.
To achieve this goal, we use an encoder that first estimates the translation $\tilde{T}$ using translation-equivariant VNT layers, then switches to a translation-invariant representation from which the rotation $\tilde{R}$ is estimated using rotation-equivariant VNN layers. Finally, the representation is made rotation-invariant and the shape encoding $\bold{Z_s}$ is generated. A reconstruction loss is computed by decoding $\bold{Z_s}$ into the canonically-positioned shape $\tilde{\bold{S}}$ and applying the extracted rigid transformation.
In the following we further explain our encoder architecture and the type of decoders used.

%3D Auto-encoding, by definition, is an equivariant operation to rotation and translation of the input point clouds. 
%Motivated by the success of folding procedure on point cloud for auto-encoding \cite{yang2018foldingnet,groueix2018papier,deprelle2019learning},
%\dcc{maybe it is better to refer to the figure up front}

\subsubsection{\SE-equivariant Encoder}
Our encoder is composed of rotation and translation equivariant and invariant layers as shown in \figref{fig:enc_arch}.
We start by feeding $\bold{X}$ through linear and non-linear VNT layers
yielding $\bold{X_{RT}} \in \Rspace^{N \times C \times 3}$, where the $RT$ subscript indicates \SE-equivariant features, as described in Section~\ref{subsec:VNT}. 
 
$\bold{X_{RT}}$ is then fed-forward through additional VNT layers resulting in a single vector neuron per point $\bold{\bar{X}_{RT}} \in \Rspace^{N \times 1 \times 3}$.  
We mean-pool the features to produce a 3D SE(3)-equivariant vector as our translation estimation, as shown in the upper branch of \figref{fig:enc_arch}, yielding $ \tilde{T} = f_T(\bold{X}) = f_T(\bold{S})R + T \in \Rspace^{1 \times 3}$, 
%\begin{equation} 
 % \tilde{T} = f_T(\bold{X}) = f_T(\bold{S})R + T \in \Rspace^{1 \times 3},
  %\label{eq:T}
%\end{equation}
where we denote by $f_T: \Rspace^{N \times 3} \rightarrow \Rspace^{1 \times 3}$ the aggregation of the VNT-layers from the input point cloud $\bold{X}$ to the estimated $\tilde{T}$, thus, it is a translation and rotation equivariant network.
 
In addition,  as explained in Section~\ref{subsec:VNT}, the following creates translation invariant features, $\bold{Y_R} = \bold{X_{RT}} - \bold{\bar{X}_{RT}} \in \Rspace^{N \times C \times 3}$.
%\begin{equation} 
%\bold{Y_R} = \bold{X_R} - \bold{\bar{X}_R} \in \Rspace^{N \times C \times 3}.
%\end{equation} 

While $\bold{Y_R}$ is translation invariant, it is still rotation equivariant, thus, we can proceed to further process $\bold{Y_R}$ with VNN layers, resulting in (deeper) rotation equivariant features $\bold{Z_R} \in \Rspace^{N \times C' \times 3}$.

Finally, $\bold{Z_R}$ is fed forward through a VNN rotation-invariant layer as explained in Section~\ref{sec:VNN}, resulting in a shape encoding, $\bold{Z_s}$, which is by construction pose invariant. Similar to the translation reconstruction, the rotation is estimated by mean pooling $\bold{Z_R}$ and feeding it through a single VN linear layer yielding $ \tilde{R} = f_R(\bold{X}) = f_R(\bold{S})R \in \Rspace^{3 \times 3},$

%\begin{equation} 
 %\tilde{R} = f_R(\bold{X}) = f_R(\bold{S})R \in \Rspace^{3 \times 3},
 %\label{eq:R}
%\end{equation}
where $f_R: \Rspace^{N \times 3} \rightarrow \Rspace^{3 \times 3}$ denotes the aggregation of the layers from the input point cloud $\bold{X}$ to the estimated rotation $\tilde{R}$ and, as such, it is a rotation-equivariant network. 
The entire encoder architecture is shown in \figref{fig:enc_arch} and we refer the reader to our supplementary for a detailed description of the layers.

\subsubsection{Decoder}
The decoder is applied on the shape encoding $\bold{Z_s}$ to reconstruct the shape $\tilde{\bold{S}}$. We stress again that $\tilde{\bold{S}}$ is invariant to the input pose, regardless of the training process.
Motivated by the success of folding networks~\cite{yang2018foldingnet,groueix2018papier,deprelle2019learning} for point clouds auto-encoding, we opt to use AtlasNetV2~\cite{deprelle2019learning} as our decoder, specifically using the point translation learning module. 
For implicit function reconstruction, we follow Occupancy network decoder~\cite{mescheder2019occupancy}.
Please note, that our method is not coupled with any decoder structure.

\subsection{Optimizing for shape-pose disentanglement}
While our auto-encoder is pose-invariant by construction, the encoding has no explicit relation to the input geometry. In the following we detail our losses to encourage a rigid relation between $\tilde{\bold{S}}$ and $X$, and for making $\tilde{\bold{S}}$ consistent across different objects. 

\subsubsection{Rigidity}
To train the reconstructed shape $\bold{\tilde{S}}$ to be isometric to the input point cloud $\bold{X}$, we enforce a rigid transformation between the two, namely ${\bold{X} = \bold{\tilde{S}}\tilde{R}+\mathbbm{1}_N\tilde{T}}$.

For point clouds auto-encoding we have used the Chamfer Distance (CD):
\begin{equation}
	\Loss_{\textit{rec}} = \textit{CD}\left(\bold{X}, \bold{\tilde{S}}\tilde{R}+\mathbbm{1}_N\tilde{T}\right),
\end{equation}
Please note that other tasks such as implicit function reconstruction use equivalent terms, as we detail in our supplementary files.

In addition, while $\tilde{R}=f_R(\bold{X})$ is rotation-equivariant we need to constraint it to SO(3), and we do so by adding an orthonormal term:
 \begin{equation}
	\Loss_{\textit{ortho}} =  \| I - \tilde{R}\tilde{R}^T\|^2_2 + \|I - \tilde{R}^T\tilde{R}\|_2^2,
\end{equation}
where $\| \cdot \|_2$ is mean square error (MSE) loss.

\subsubsection{Consistency}
Now, our shape reconstruction $\bold{\tilde{S}}$ is isometric to $\bold{X}$ and it is invariant to $T$ and $R$. However, there is no guarantee that the pose of $\bold{\tilde{S}}$ would be consistent across different instances. 

\begin{figure*}[t]
	\centering
	\includegraphics[width=1.0\linewidth]{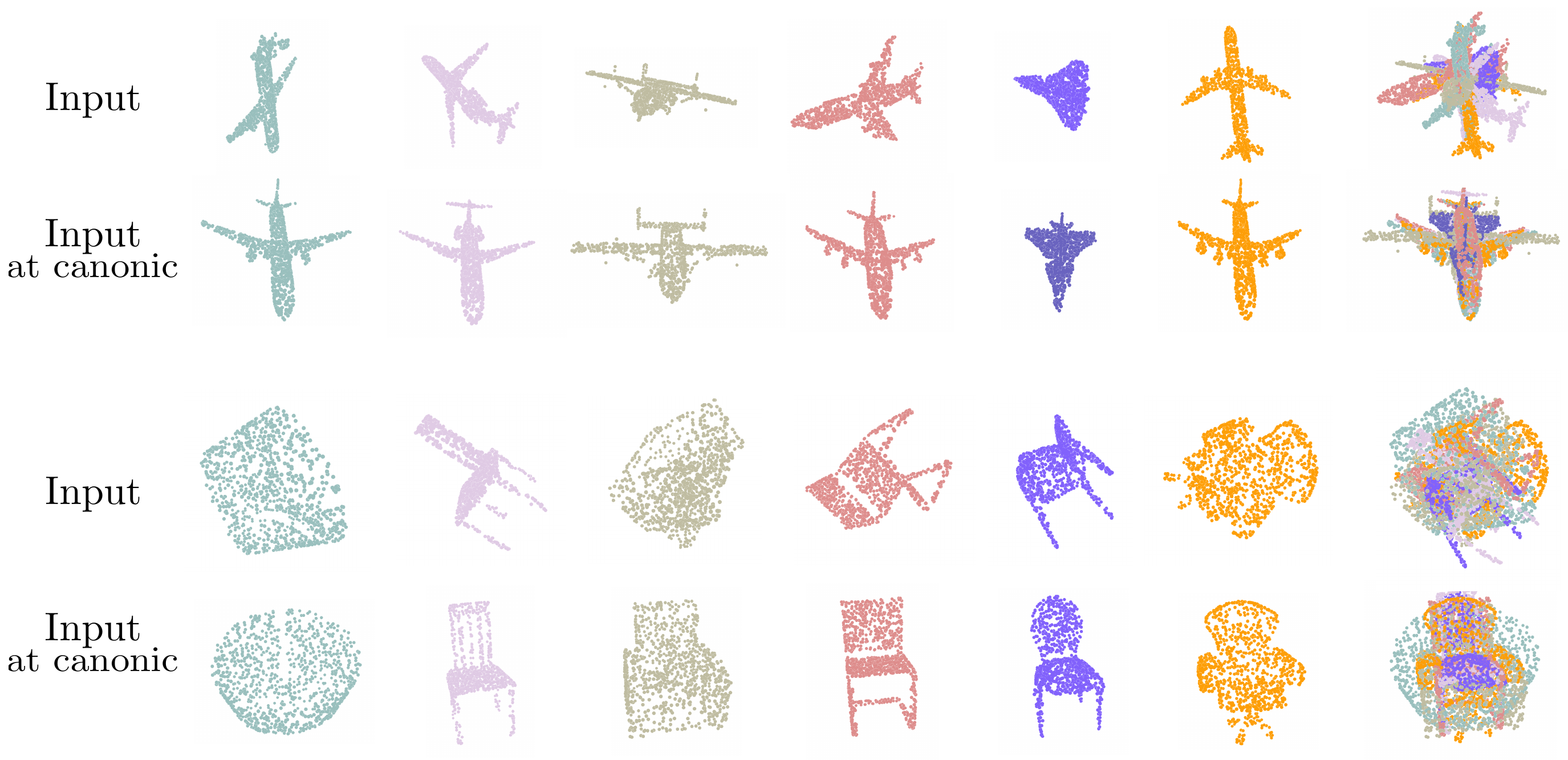}
	\caption{Aligning planes and chairs. The input planes and chairs (first and third row, respectively) have different shapes and different poses, as can be seen separately and together (rightmost column). We apply the inverse learned pose, transforming the input to its canonical pose (second and fourth row). }
	\label{fig:consistency}
\end{figure*}

Assume two different point clouds $\bold{X_1},\bold{X_2}$ are aligned. If their canonical representations $\bold{S_1},\bold{S_2}$ are also aligned, then they have the same rigid transformation w.r.t their canonical representation and vice versa, i.e., $\bold{X_i} = \bold{S_i}R + \mathbbm{1}_NT$, $i=1,2$.
% \begin{align}
%	\bold{X_1} &= \bold{S_1}R + T, \\ \nonumber
%	\bold{X_2} &= \bold{S_2}R + T, 
%\end{align}
To achieve such consistency, we require: 
\begin{equation}
\{f_T(\bold{X_1}), f_R(\bold{X_1})\}= \{f_T(\bold{X_2}), f_R(\bold{X_2})\}.
\end{equation}

%Examining Eq.~\eqref{eq:T} and Eq.~\eqref{eq:R}, it is easy to see that such requirement can be achieved by
 %\begin{equation}
 %	f_T(\bold{S}) = 0, \quad \textrm{and} \quad f_R(\bold{S}) = I,
%\end{equation}
%i.e., the pose estimation should be invariant to the shape. 

%To achieve such shape invariance
We generate such pairs of aligned point clouds, by augmenting the input point cloud $\bold{X}$ with several simple augmentation processes, which do not change the pose of the object. 
In practice, we have used Gaussian noise addition, furthest point sampling (FPS), patch removal by k-nn (we select one point randomly and remove $k$ of its nearest neighbors) and re-sampling of the input point cloud.

We then require that the estimated rotation and translation is the same for the original and augmented versions,
 \begin{equation}
	\Loss^{\textit{aug}}_{\textit{consist}} = \sum_{A \in \mathcal{A}} \| f_R\left(\bold{X}\right) - f_R\left(A(\bold{X})\right)\|_2^2 +  \| f_T\left(\bold{X}\right) - f_T\left(A(\bold{X})\right)\|_2^2,
\end{equation}
where $\mathcal{A}$ is the group of pose preserving augmentations and $\|\cdot \|_2$ is MSE loss.

%\begin{figure}[t]
%	\centering
%	\includegraphics[width=1.0\linewidth]{figures/aug_example.pdf}
%	\caption{Shape augmentations. The source point cloud on the left, is modified to supervise a rotation and translation learning invariant of the shape.
%	The different augmentations are shown in brown, where the same source point cloud can be seen in each augmentation in blue
%	\label{fig:aug}
%\end{figure}

In addition, for point cloud reconstruction, we can also generate a version of $\bold{X}$, with a known pose, by feeding again the reconstructed shape $\bold{\tilde{S}}$. We transform $\bold{\tilde{S}}$ by a random rotation matrix $R^*$ and a random translation vector $T^*$
and require the estimated pose to be consistent with this transformation:
 \begin{equation}
	\Loss^{\textit{can}}_{\textit{consist}} = \| f_R\left(\bold{\tilde{S}}R^*+T^*\right) - R^*\|_2^2 +  \| f_T\left(\bold{\tilde{S}}R^*+T^*\right) - T^*\|_2^2,
\end{equation}

Our overall loss is
 \begin{equation}
	\Loss =  \Loss_{\textit{rec}} + \lambda_1\Loss_{\textit{ortho}} + \lambda_2\Loss^{\textit{aug}}_{\textit{consist}}  + \lambda_3\Loss^{\textit{can}}_{\textit{consist}},
\end{equation}
where the $\lambda_i$ are hyper parameters, whose values in all our experiments were set to $\lambda_1=0.5$, $\lambda_2=\lambda_3=1$.

%\begin{equation}
  %  X_{inv}f_R\left({X_{can}}\right)R + f_{T}\left(X_{can}\right)R + T
%\end{equation}

%Therefore, for $X_inv$ to equal $X_can$, we must regularize either $f_{pc}$ to be consistent across shapes or $f_T$, $f_R$ to be invariant to the actual shape

\subsection{Inference}
At inference time, we feed forward point cloud $\bold{X} \in \Rspace^{N \times 3}$ and retrieve its shape and pose. However, since our estimated rotation matrix $\tilde{R}$ is not guaranteed to be orthonormal, at inference time, we find the closest ortho-normal matrix to $\tilde{R}$ (i.e., minimize the Forbenius norm), following~\cite{bar1975iterative}, by solving:
\begin{equation}
	\hat{R} = \tilde{R}\left(\tilde{R}^T\tilde{R}\right)^{-\frac{1}{2}}.
\end{equation}
The inverse of the square root can be computed by singular value decomposition (SVD). While this operation is also differentiable we have found it harmful to incorporate this constraint during the training phase, thus it is only used during inference. We refer the reader to \cite{bar1975iterative} for further details.
\begin{figure*}[t]
	\centering
	\includegraphics[width=1.0\linewidth]{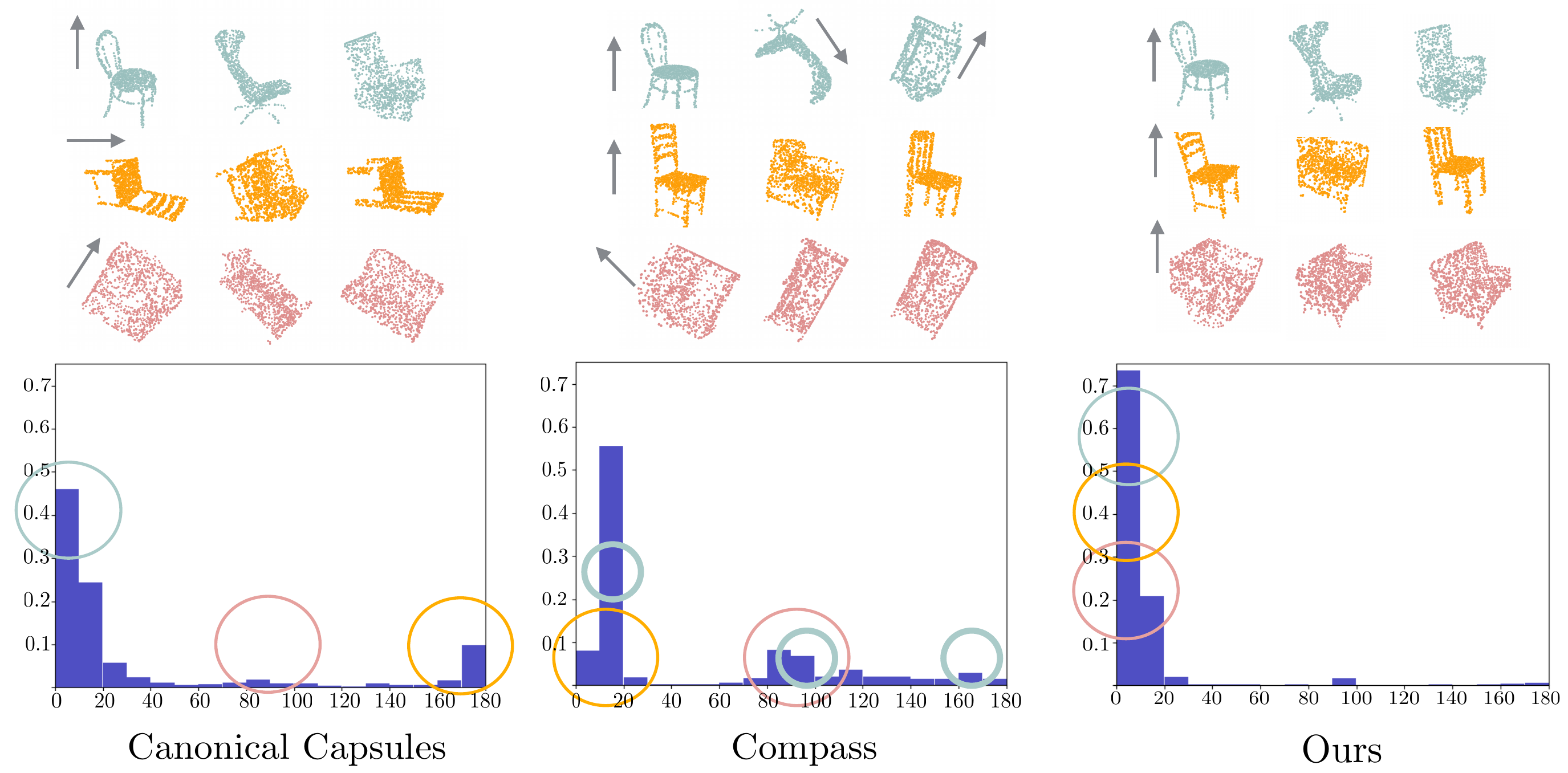}
	\caption{Histogram of canonical pose deviation from the mean canonical pose. We estimate the canonical pose of aligned 3D point clouds from ShapeNet using our method, Canonical Capsules~\cite{sun2021canonical} and Compass \cite{spezialetti2020learning}.
	In the bottom row, we show the normalized histogram of the deviation from the mean pose. It is clear that while our method is shape-consistent, both Compass and Canonical Capsules struggle to have a single canonical pose.
	In the top row, we focus on small, medium and large deviation cases of Canonical Capsules, marked by cyan, red, and orange circles on the histogram plot, respectively. The canonical pose of the same objects is shown for Compass and our method, as well as their location on the corresponding histogram plot. The arrow next to the objects is directed toward the local shape z+ direction. }
	\label{fig:consistency_hist}
\end{figure*}

\section{Results}
We preform qualitative and quantitative comparison of our method for learning shape-invariant pose. Due to page limitations, more results can be found in our supplementary files.

\subsection{Dataset and implementation details}
We employ the ShapeNet dataset~\cite{chang2015shapenet} for evaluation. For point cloud auto-encoding we follow the settings in \cite{sun2021canonical} and \cite{deprelle2019learning}, and use ShapeNet Core focusing on two categories: airplanes and chairs. %(with the train-test split as in \cite{deprelle2019learning}). 
While airplanes are more semantically consistent and containing less variation, chairs exhibit less shape-consistency and may contain different semantic parts. All 3D models are randomly rotated and translated in the range of $[-0.1,0.1]$ at train and test time. 

 For all experiments, unless stated otherwise, we sample random $1024$ points for each point cloud. The auto-encoder is trained using Adam optimizer with learning rate of $1e^{-3}$ for $500$ epochs, with drop to the learning rate at $250$ and $350$ by a factor of $10$. We save the last iteration checkpoint and use it for our evaluation. The decoder is AtlasNetV2~\cite{deprelle2019learning} decoder with $10$ learnable grids. 
 
\subsection{Pose consistency}
We first qualitatively evaluate the consistency of our canonical representation as shown in \figref{fig:consistency}. 
At test time, we feed different instances at different poses through our trained network, yielding estimated pose of the input object w.r.t the pose-invariant shape. We then apply the inverse transformation learned, to transform the input to its canonical pose. As can be seen, the different instances are roughly aligned, despite having different shapes.
 More examples can be found in our supplementary files.

We also compare our method, both qualitatively and quantitatively, to Canonical Capsules~\cite{sun2021canonical} and Compass~\cite{spezialetti2020learning} by using the alignment in ShapeNet (for Compass no translation is applied). First, we feed forward all of the aligned test point clouds $\left\{\bold{X}_i \right\}_{i=1}^{N_t}$ through all methods and estimate their canonical pose $\{\tilde{R}_i\}_{i=1}^{N_t}$.  We expect to have a consistent pose for all aligned input shapes, thus, we quantify for each instance $i$ the angular deviation $d^{\textit{consist}}_i$ of its estimated pose $\tilde{R}_i$ from the mean pose $d^{\textit{consist}}_i = \angle{\left(\tilde{R}_i, \frac{1}{N_t}\sum_i \tilde{R}_i\right)}.$
%\begin{equation}
%	d^{\textit{consist}}_i = \angle{\left(\tilde{R}_i, \frac{1}{N_t}%\sum_i \tilde{R}_i\right)}.
%\end{equation}
We present an histogram of $\{d^{\textit{consist}}_i\}_{i=1}^{N_t} $ in \figref{fig:consistency_hist}.
As can be seen, our method results in a more aligned canonical shapes as indicated by the peak around the lower deviation values. We visualize the misalignment of Canonical Capsules by sampling objects with small, medium and large deviation, and compare them to the canonical representation achieved  by Compass and our method for the same instances.
The misalignment of Canonical Capsules may be attributed to the complexity of matching unsupervised semantic parts between chairs as they exhibit high variation (size, missing parts, varied structure). 
We quantify the consistency by the standard deviation of the estimated pose $\sqrt{\frac{1}{N_t}\sum_i{d_i^2}}$  in Table~\ref{table:stable_consist}. Evidently, Compass falls short for both object classes. Canonical Capsules preform slightly better than our method for planes, while our method is much more consistent for the chair category. 

\begin{figure*}[t]
	\centering
	\includegraphics[width=1.0\linewidth]{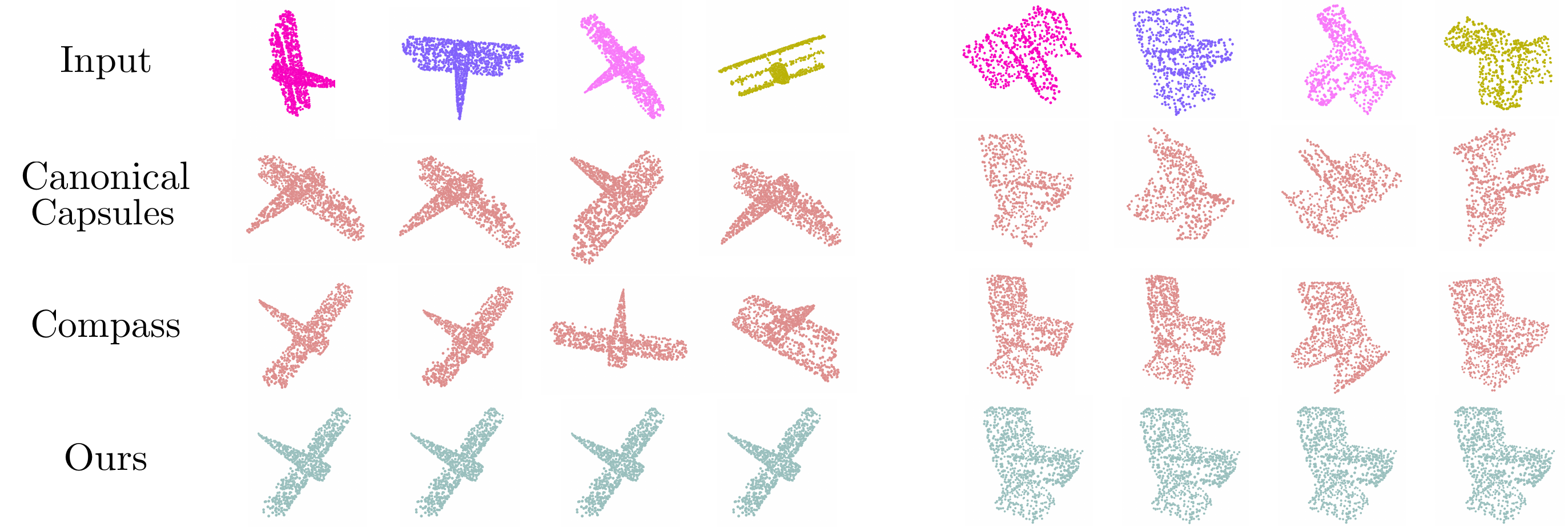}
	\caption{Stability of the canonical representation to rigid transformation of the input. The location and orientation of the same point cloud affects its canonical representation in both  Canonical Capsules~\cite{sun2021canonical} and Compass~\cite{spezialetti2020learning}. Our canonical representation (bottom row) is \SE-invariant to the rigid transformation of the input.}
	\label{fig:stability}
\end{figure*}
\begin{figure*}
	\centering
	\includegraphics[width=1.0\linewidth]{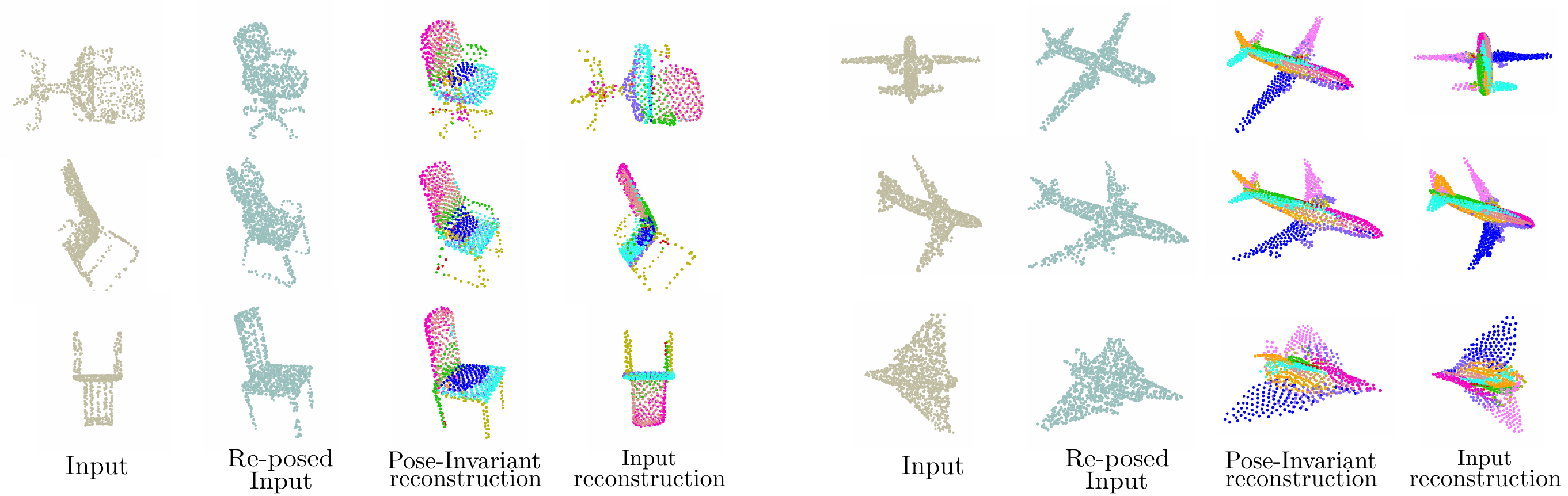}
	\caption{Reconstruction of chairs and planes under SE(3) transformations. The input point cloud (left) is disentangled to shape (second from the right) and pose, which together reconstruct the input point cloud, as shown in the right most column. The inverse pose is applied to the input point cloud to achieve a canonical representation (second image from the left). The colors of the reconstructed point cloud indicate different decoders of AtlasNetV2~\cite{deprelle2019learning}}. 
	\label{fig:recon_ae}
	\vspace{-4mm}
\end{figure*}

\setlength{\tabcolsep}{4pt}
\begin{table}
\begin{center}
\caption{Stability and consistency of the estimated pose, lower is better.}
\label{table:stable_consist}
\begin{tabular}{ l c c c | c c c}
\toprule
\multicolumn{2}{c}{$$} & \multicolumn{1}{c}{Stability} & \multicolumn{2}{c}{$$} & \multicolumn{1}{c}{Consistency} \\
\cmidrule(lr){2-4}
\cmidrule(lr){5-7} 
\multicolumn{1}{c}{$$}  & \small{Capsules} & \small{Compass} & \small{Ours}  & \small{Capsules} & \small{Compass} & \small{Ours} \\
\midrule
\small{Airplanes}
& 7.42 & 13.81 & $\bold{2e^{-3}}$ & $\bold{45.76}$ & $71.43$ & $ 49.97$  \\
\small{Chairs}
& 4.79 & 12.01  & $\bold{4e^{-3}}$ & $68.13$ & $68.2$ & $\bold{24.31}$ \\ %$
%%\small{Multiple} &  18.98 & -- & \textbf{--} &  -- & -- &  \textbf{--} \\
\hline
\end{tabular}
\end{center}
\end{table}
\setlength{\tabcolsep}{1.4pt}
\subsection{Stability}
A key attribute in our approach is the network construction, which outputs a purely \SE-invariant canonical shape. 
Since we do not require any optimization for such invariance, our canonical shape is expected to be very stable compared with Canonical Capsules and Compass.
We quantify the stability, as proposed by Canonical Capsules, in a similar manner to the consistency metric. For each instance $i$, we randomly rotate the object $k=10$ times, and estimate the  canonical pose for each rotated instance $\{\tilde{R}_{ij}\}_{j=1}^k$. We average across all $N_t$ instances the standard deviation of the angular pose estimation as follows,
\begin{equation}
d^{\textit{stability}} = \frac{1}{N_t}\sum_i{\sqrt{\sum_j\angle{\left(\tilde{R}_{ij}, \frac{1}{k}\sum_j \tilde{R}_{ij}\right)}^2 }}.
\end{equation}

The results are reported in Table~\ref{table:stable_consist}. 
%As expected, the optimization based invariance is not stable for all instances as it relies on the network to generalize both to different shapes and different poses. 
As expected, Canonical Capsules and Compass exhibit non-negligible instability, as we visualize in \figref{fig:stability}.  

\begin{figure*}[t]
	\centering
	\includegraphics[width=1.0\linewidth]{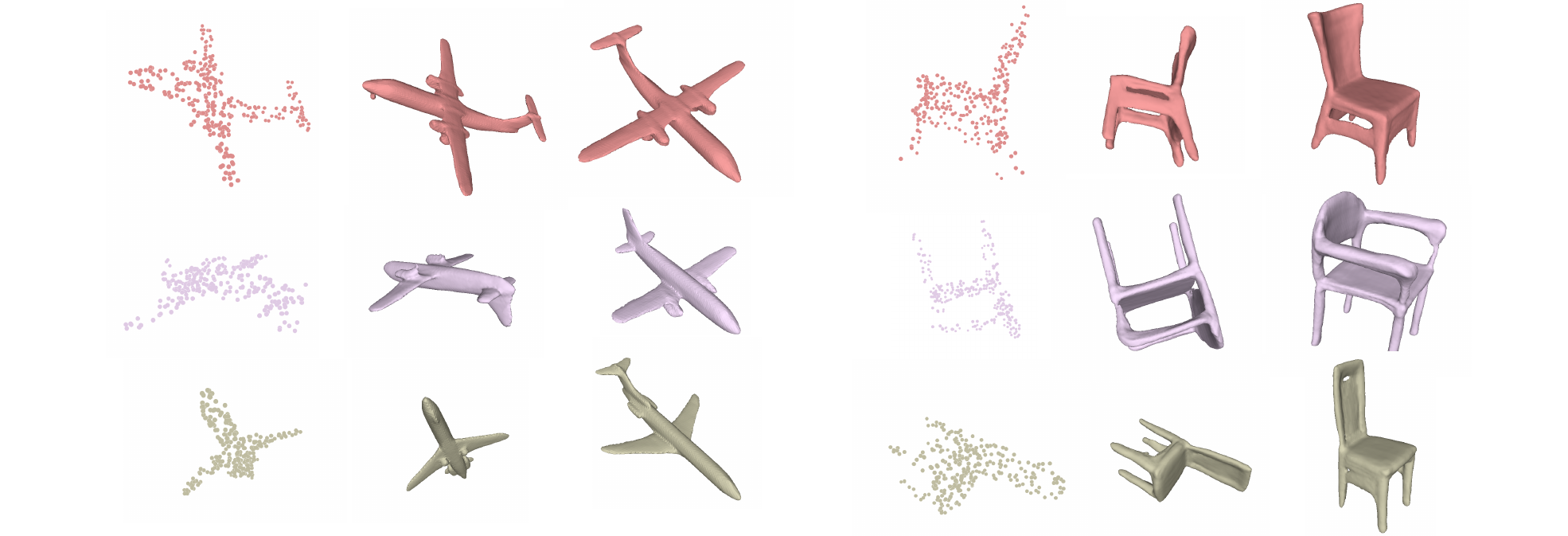}
	\caption{Reconstruction results of OccNet~\cite{mescheder2019occupancy} via shape-pose disentanglement. An input point cloud on the left is disentangled to shape encoding and pose. OccNet decodes only the shape encoding yielding a canonical shape on the right column. The reconstruction is then transformed by the estimated pose as seen in the middle column. Meshes are extracted via Multiresolution IsoSurface Extraction (MISE)~\cite{mescheder2019occupancy} }
	\label{fig:implicit}
\end{figure*}

\subsection{Reconstruction quality}
We show qualitatively our point cloud reconstruction in \figref{fig:recon_ae}. Please note that our goal is not to build a SOTA auto-encoder in terms of reconstruction, rather we learn to disentangle pose from shape via auto-encoding. Nonetheless, our auto-encoder does result in a pleasing result as shown in \figref{fig:recon_ae}. Moreover, since we utilize AtlasNetV2\cite{deprelle2019learning} which utilizes a multiple patch-based decoder, we can examine which point belongs to which decoder. As our shape-encoding is both invariant to pose and consistent across different shapes, much like in the aligned scenario, each decoder assume some-what of semantic meaning, capturing for example the right wing of the airplanes. Please note that we do not enforce any structuring on the decoders. 

%In this section we ablate our augmentation losses and report the reconstruction loss and the consistency loss in Tab.~\ref{tab:ablation}.
%Evidently, all three losses are important to achieve consistent canonical representation.

\subsection{3D implicit reconstruction}
We show that our encoder can be attached to a different reconstruction task by repeating OccNet~\cite{mescheder2019occupancy} completion experiment. We replace OccNet encoder with our shape-pose disentagling encoder. The experiment is preformed with the same settings as in \cite{mescheder2019occupancy}. We use the subset of \cite{choy20163d}, and the point clouds are sub-sampled from the watertight mesh, containing only $300$ points and applied with a Gaussian noise. We have trained OccNet for $600K$ iterations and report the results of the best (reconstruction wise) checkpoint.
%Overall we train two occupancy networks for chairs and planes and  reconstruct a pose-invariant occupancy function $o: \Rspace^3 \rightarrow {0,1}$. 
We show in \figref{fig:implicit} a few examples of rotated point clouds (left), its implicit function reconstruction (middle) and the implicit function reconstruction in the canonical pose (right).

%\subsection{Robstness to occlusion and noise}
%Our augmentations technique directly utilize occlusion and noise to supervise the learned shape to be consistent. We compare the robustness of our approach to canonical capsules and compass, where we have added similar augmentations to both at training time. As shown in Fig.\ref{fig: }

\section{Conclusions}
We have presented a stable and consistent canonical representation learning. To achieve a pose-invariant represenation, we have devised an \SE-equivairant encoder, extending the VNN framework, to meet the requirements of canonical pose learning, i.e., learning rigid transformations. Our experiments show, both qualitatively and quantitatively, that our canonical representation is significantly more stable than recent approaches and has similar or better consistency, especially for diverse object classes. 
Moreover, we show that our approach is not limited to specific decoding mechanism, allowing for example to reconstruct canonical implicit neural field.
%Hence, we believe that our canonical representation can be useful in tasks requiring aligned settings such as completion and unsupervised segmentation, not only as a pre-alignment tool, but it can be learned simultaneously.
In the future, we would like to explore the potential of our canonical representation for point cloud processing tasks requiring aligned settings, such as completion and unsupervised segmentation, where the canonical representation is learned on-the-fly, along with the task.

\newpage
\bibliographystyle{ACM-Reference-Format}
\bibliography{egbib}
\clearpage
\newpage
\appendix
\section{\SE-equivariance verification}
In this section we verify that our VNT layers are indeed translation and rotation equivariant, as well as explicitly present other layers, not included in the paper.

\subsection{Verifying \SE-equivariant linear layer} 
We verify that the linear module $f_{\textrm{lin}}\left(\cdot;\bold{W}\right)$, defined via a weight matrix \\ $\bold{W} \in \mathcal{W}^{C' \times C}$, acting on a vector-list feature $\bold{V} \in \Rspace^{C \times 3}$, such that
\begin{equation}
\mathcal{W}^{C' \times C} = \left\{\bold{W} \in \Rspace^{C' \times C} \mid \textstyle{\sum_{j=1}^{C}} w_{i,j} = 1 \quad \forall i \in [1,C'] \right\},
\end{equation}
is \SE-equivariant.

Let $\bold{w_j} \in \Rspace^{C' \times C}$ be the $j$ column of $\bold{W}$,  and let $R \in \Rspace^{3 \times 3}$ be a rotation matrix and $T \in \Rspace^{1\times 3}$ a translation vector.
For  $f_{\textrm{lin}}\left(\cdot;\bold{W}\right)$  to be \SE-equivariant, the following must hold:
\begin{align}
   &  f_{\textrm{lin}}\left(\bold{V}R + \mathbbm{1}_CT\right) = \bold{W}\left(\bold{V}R + \bold{\mathbbm{1}}_{C}T\right) = \bold{W}\bold{V}R + \bold{W}\bold{\mathbbm{1}}_{C} T = \nonumber \\
     & = \bold{W}\bold{V}R + \left(\sum_{j=1}^C \bold{w}_j\right)T =
      (\bold{W}\bold{V})R + \bold{\mathbbm{1}}_{C'} T,  =  f_{\textrm{lin}}(\bold{V})R + \bold{\mathbbm{1}}_{C'} T, 
    \label{eq:lin_eq_T}
\end{align}
where $\mathbbm{1}_C = [1,1,\dots, 1]^T \in \Rspace^{C \times 1}$ is a column vector of length $C$, and $\sum_{j=1}^C \bold{w_j} = \mathbbm{1}_{C'}$, since $\left(\sum_{j=1}^C \bold{w_j}\right)[i] = \sum_{j=1}^C {w_{ij}} = 1 $ for $i=1,\dots, C'$

\subsection{Verifying \SE-equivariant ReLU}
We verify that the ReLU layer is \SE-equivariant.

Let  $\bold{V}, \bold{V'} \in \Rspace^{C \times 3}$ be the input and output of a ReLU layer,
\begin{equation}
	\bold{V'} = f_{\textrm{ReLU}}\left(\bold{V}\right).
\end{equation}
Let  ${\bold{v'} \in  \Rspace^{1 \times 3}}$ be a single vector, such that $\bold{v'} \in \bold{V'}$.
As explained in Section 3.1 of the paper, we learn three translation equivariant linear maps,  $\bold{Q}, \bold{K}, \bold{O} \in \mathcal{W}^{1 \times C}$ projecting the input to $\bold{q}, \bold{k}, \bold{o} \in \Rspace^{1 \times 3}$, yielding an origin $\bold{o}$, a feature $\bold{q_o} = \bold{q} - \bold{o}$ and a direction  $\bold{k_o} = \bold{k} - \bold{o}$.
The ReLU layer for a single vector neuron is then defined via
\begin{equation}
  \bold{v'} =\begin{cases}
    \bold{o} + \bold{q_o} & \text{if} \quad \innerprod{\bold{q_o}}{\bold{k_o}} \geq 0, \\
    \bold{o} + \bold{q_o} - \innerprod{\bold{q_o}}{\frac{\bold{k_o}}{\norm{\bold{k_o}}}}\frac{\bold{k_o}}{\norm{\bold{k_o}}} , & \text{otherwise}.
  \end{cases}.
\end{equation}

$\bold{q}, \bold{k}, \bold{o}$ are \SE-equivariant and according to Eq .(7) of the paper,  $\bold{q_o}, \bold{k_o}$ are translation invariant (and rotation equivariant) as they are the subtraction of two \SE-equivariant vector neurons, thus, the condition term $\innerprod{\bold{q_o}}{\bold{k_o}}$ is also translation invariant.
As shown in VNN~[6], the inner product of two rotation equivariant vector-neurons is rotation invariance. Similarly here, assume the input $\bold{V}$ is rotated with a rotation matrix $R \in \Rspace^{3 \times 3}$, then
\begin{equation}
\innerprod{\bold{q_o}R}{\bold{k_o}R} = \bold{q_o}RR^T\bold{k_o} = \bold{q_o}\bold{k_o}^T = \innerprod{\bold{q_o}}{\bold{k_o}} 
\label{eq:inv_R}
\end{equation}

To conclude, the condition term $\innerprod{\bold{q_o}} {\bold{k_o}}$ is \SE-invariant.

When $\innerprod{\bold{q_o}} {\bold{k_o}} \geq 0$, the output vector neuron $\bold{v'} = \bold{o} + \bold{q_o} = \bold{q}$, and thus it is \SE-equivariant.

When $\innerprod{\bold{q_o}} {\bold{k_o}} < 0$ the output vector neuron is
\begin{equation}
\bold{o} + \bold{q_o} - \innerprod{\bold{q_o}}{\frac{\bold{k_o}}{\norm{\bold{k_o}}}}\frac{\bold{k_o}}{\norm{\bold{k_o}}},
\end{equation}

Similarly to Eq~.\eqref{eq:inv_R}{} the term 
\begin{equation}
\innerprod{\bold{q_o}}{\frac{\bold{k_o}}{\norm{\bold{k_o}}}}\frac{1}{\norm{\bold{k_o}}}  = \frac{\innerprod{\bold{q_o}}{\bold{k_o}}}{\innerprod{\bold{k_o}}{\bold{k_o}}},
\end{equation}
is also \SE-invariant.

We can now easily prove that if the input $\bold{V}$ is rotated with a rotation matrix $R \in \Rspace^{3 \times 3}$ and translation vector $T \in \Rspace^{1 \times 3}$, then

\begin{align}
  \bold{v'} = &\bold{o}R +\mathbbm{1}_{C}T + \bold{q_o}R - \innerprod{\bold{q_o}}{\frac{\bold{k_o}}{\norm{\bold{k_o}}}}\frac{\bold{k_o}R}{\norm{\bold{k_o}}} \nonumber \\ & = \left(\bold{o} + \bold{q_o} - \innerprod{\bold{q_o}}{\frac{\bold{k_o}}{\norm{\bold{k_o}}}}\frac{\bold{k_o}}{\norm{\bold{k_o}}}\right)R + \mathbbm{1}_{C}T 
  = \bold{v'}R + \mathbbm{1}_{C}T,
\end{align}
Thereby completing the proof.

\subsection{VNT-LeakyReLU}
%Neglecting the linear mapping of the input to feature $\bold{q}$ The LeakyReLU can be expressed as
%\begin{equation}
%\bold{V'} = f_{\textrm{LeakyReLU}}\left(\bold{V} ;\alpha \right) = \alpha \bold{V} + (1-\alpha)f_{\textrm{ReLU}}\left(\bold{V}\right)
%\end{equation}
LeakyReLU is defined in a similar manner to the ReLU layer, with slight modification to the output vector neuron, given by
\begin{equation}
\bold{v'} = \alpha \bold{q} + \left(1 - \alpha\right) \bold{v'}_{\textrm{ReLU}},
\end{equation}
 where $\alpha \in \Rspace$

Easy to see that the $\bold{v'}$ is \SE-equivariant.

\subsection{VNT-MaxPool}
Given a set of vector-neuron list $\mathcal{V} \in \Rspace^{N \times C \times 3}$, we learn two linear maps $\bold{K},\bold{O} \in \mathcal{W}^{C \times C}$, shared between $\bold{V_n} \in \mathcal{V}$.

We obtain a translation invariant direction
\begin{equation}
	\mathcal{K} = \left\{\bold{K}\bold{V_n} -\bold{O}\bold{V_n},\right\}_{n=1}^N
\end{equation}
and a translation invariant features
\begin{equation}
	\mathcal{Q} = \left\{\bold{V_n} - \bold{O}\bold{V_n}\right\}_{n=1}^N.
\end{equation}

The VNT-MaxPool is defined by
\begin{align}
	& f_{MAX}\left(\mathcal{V}\right)[c] = \bold{V}_{n^*}[c] \\
	& \textrm{where} \quad n^*=  \underset{n}{\argmax} \innerprod{\bold{Q_n}[c]}{\bold{K_n}[c]},
\end{align}
where $\bold{Q_n} \in \mathcal{Q}$ and $\bold{K_n} \in \mathcal{K}$.
Since $\bold{Q_n}$, $\bold{K_n}$ are translation invariant, and their inner product is also rotation invariant the selection process of $ n^*$ for every channel $c$ is invariant to \SE.
We note that both $\bold{K},\bold{O}$ can be shared across vector-neurons.

\section{Implementations Details}
\subsection{Encoder architecture}
In this section we elaborate on our encoder architecture. Our encoder contains VNT layers following with VNN layers as reported in Table~\ref{table:enc}. LinearLeakyReLU stands for the leakyReLU with feature learning $\mathcal{Q}$. For the exact VNN layers definition (and specifically STNkd) we refer the reader to VNN~[6].
\setlength{\tabcolsep}{4pt}
\begin{table}
\begin{center}
\caption{Our shape-pose disentangling encoder architecture. }
\label{table:enc}
\begin{tabular}{  c | c c c c}
\toprule

Name & Input channel & Output channel & Type \\ 
\midrule
\small{LinearLeakyReLU} & $3$ & $64$//$3$ &  \small{VNT}  \\ %$
\small{LinearLeakyReLU} & $64$//$3$ & 64//3 &  \small{VNT}   \\ %$
\small{T-invariant} & $64$//$3$ & $64$//$3$ &  \small{VNT}  \\
\small{LinearLeakyReLU} & $64$//$3$ & $64$//$3$ &  \small{VNN}  \\
\small{STNkd+concat} & $64$//$3$ & $2\cdot$($64$//$3$) &  \small{VNN}  \\
\small{LinearLeakyReLU} &  $2\cdot$($64$//$3$) & $2\cdot$($64$//$3$) &  \small{VNN}  \\
\small{LinearLeakyReLU} & $2\cdot$($64$//$3$) &  $170$ &  \small{VNN}  \\
\small{BatchNorm} & $170$ & $170$  &  \small{VNN} \\
\small{Meanpool+concat} &  $170$ & $340$ &  \small{VNN}  \\
\small{R-invariant} &  $340$ &  $340$ &  \small{VNN}   \\
\small{Flatten} &   $340\cdot 3$ &  $1020$ &  \small{Regular}   \\
\small{Max-pool} &   $1020$ &  $1020$ &  \small{Regular}   \\
%%\small{Multiple} &  18.98 & -- & \textbf{--} &  -- & -- &  \textbf{--} \\
\hline
\end{tabular}
\end{center}
\end{table}
\setlength{\tabcolsep}{1.4pt}

%\small{LinearLeakyReLU} &  340 & 170 &  \small{VNN} & \\
%\small{LinearLeakyReLU} &  170 & 85 &  \small{VNN} &  \\
%\small{LinearLeakyReLU} &  85 & 3 &  \small{VNN} &  \\

\section{Implicit reconstruction}
Occupancy network reconstruction from a point clouds $X \in \Rspace^{N \times 3}$, with a learned embedding $\bold{z} \in \mathcal{X}$ , learns a mapping function $f_{\theta}(p, \bold{z}): \Rspace^3 \times \mathcal{X} \rightarrow [0,1]$.
In occupancy network completion experiment, the point cloud is sampled from the watertight mesh, and the mesh is used as supervision to sample $M$ training point $\left\{p_i\right\}_{i=1}^M$ inside and outside the mesh, indicated by $\left\{o_i\right\} \in [0,1]^M$. We follow the same experiment, with slight changes. We feed our learned pose-invariant encoding $\bold{Z_s}$ through  $f_{\theta}$, and project the points $\left\{p_i\right\}$ from the input pose to the learned canonical pose by:
\begin{equation}
	\tilde{p}_i = (p_i - \tilde{T})\tilde{R}^T \quad i=1,\dots,M
\end{equation}
Therefore, our reconstruction loss is
\begin{equation}
	\Loss_{\textit{rec}} = \sum_{i=1}^M{\mathcal{L}_{\textrm{BCE}}\left(f_{\theta}(\tilde{p}_i, \bold{Z_s}), o_i \right)} ,
\end{equation}
where $\mathcal{L}_{\textrm{BCE}}$ is binary cross entropy loss.
For implicit reconstruction we have found it beneficial to train the network in an alternating approach, where at the first phase we backward w.r.t $\Loss_{\textit{rec}}$ and in the second phase we backward w.r.t 
 \begin{equation}
	\Loss_2 =  \lambda_1\Loss_{\textit{ortho}} + \lambda_2\Loss^{\textit{aug}}_{\textit{consist}}.
\end{equation}

\section{Additional results}
\subsection{Augmentations ablation}
\begin{figure}[t]
	\centering
	\includegraphics[width=1.0\linewidth]{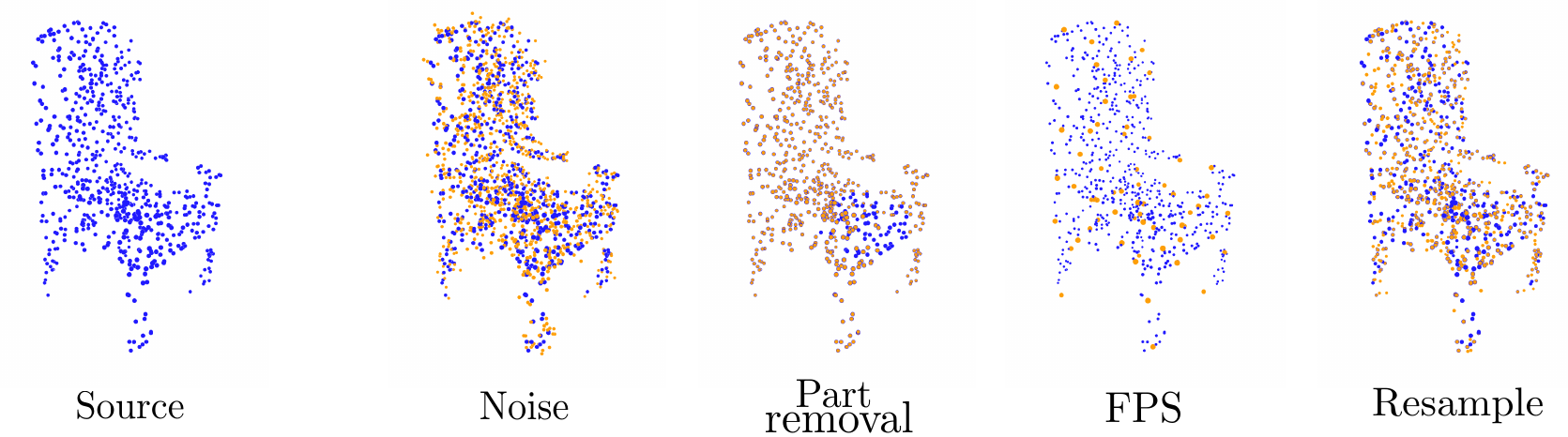}
	\caption{ Point cloud augmentations. The source point cloud on the left (blue), is modified (brown) to supervise a rotation and translation learning invariant of the shape. }
	\label{fig:aug}
\end{figure}

In this section we specify in more details our augmentations, which can be seen in \figref{fig:aug}, and ablate their individual donation to the consistency of our canonical representation. Our augmentations are Furthest point sampling (FPS), with random number of points $N_{\textrm{FPS}} = \textrm{U}\left(300, 500\right)$ per batch, K-NN removal (KNN), where a point is randomly selected on the point cloud, and its $N_{\textrm{KNN}}=100$ points are removed, Gaussian Noise added to the point clouds with $\mu=0$ and $\sigma=0.025$, a re-sampling augmentations (Resample) where we re-select which $N=1024$ to sample from the original point cloud, and canonical rotation (Can), where the point cloud reconstruction in its canonical representation is rotated and transformed to create a supervised version of itself.
Since our method is pose-invariant by construction, different augmentations have no effect on the stability, thus, we ablate only w.r.t the consistency as reported in Table~\ref{table:ablation}. 
 We ablate the donation of each augmentation by removing it from the training process and measuring the consistency as defined in the paper. Evidently, the lesser factor is the noise addition augmentation, while KNN and FPS donate the most to the consistency metric.

\setlength{\tabcolsep}{4pt}
\begin{table}
\begin{center}
\caption{Consistency of different augmentations composition (lower is better). Each column but the last represents the absence of an augmentation, indicating its importance to the consistency metric.}
\label{table:ablation}
\begin{tabular}{  c | c c c c c | c | c}
\toprule

\multicolumn{1}{c}{$$} & \small{-FPS} & \small{- Noise} & \small{-KNN}  &  \small{-Resample} & \small{-Can}  & All\\
\midrule
\small{Airplanes}
& 52.31 & 50.0& $66.4$ & $50.5$ & $53.7$ & $ {\bf 49.9}$  \\
\small{Chairs}
& 27.31 & 24.41  & $27.31$ & $25.3$ & $25.1$ & $\bold{24.31}$ \\ %$
%%\small{Multiple} &  18.98 & -- & \textbf{--} &  -- & -- &  \textbf{--} \\
\hline
\end{tabular}
\end{center}
\end{table}
\setlength{\tabcolsep}{1.4pt}

\subsection{Pose consistency}
\begin{figure*}[t]
	\centering
	\includegraphics[width=1.0\linewidth]{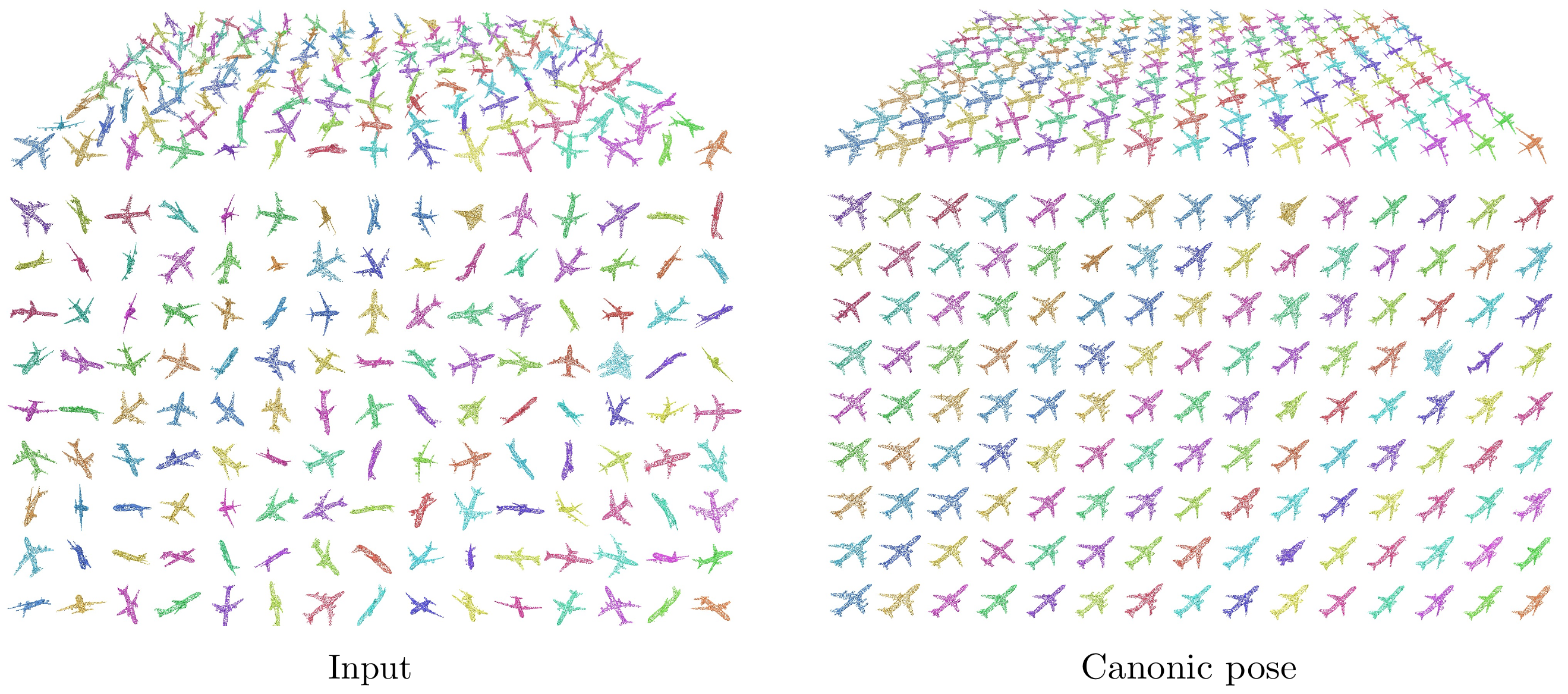}
	\caption{More results of planes in canonic representations. The planes on the left are randomly translated and rotated, as seen from a side view (first row) and top view (second row). Our canonical representation, on the right, exhibit good alignment across different instances both in orientation and position.}
	\label{fig:planes_can}
\end{figure*}

\begin{figure*}[t]
	\centering
	\includegraphics[width=1.0\linewidth]{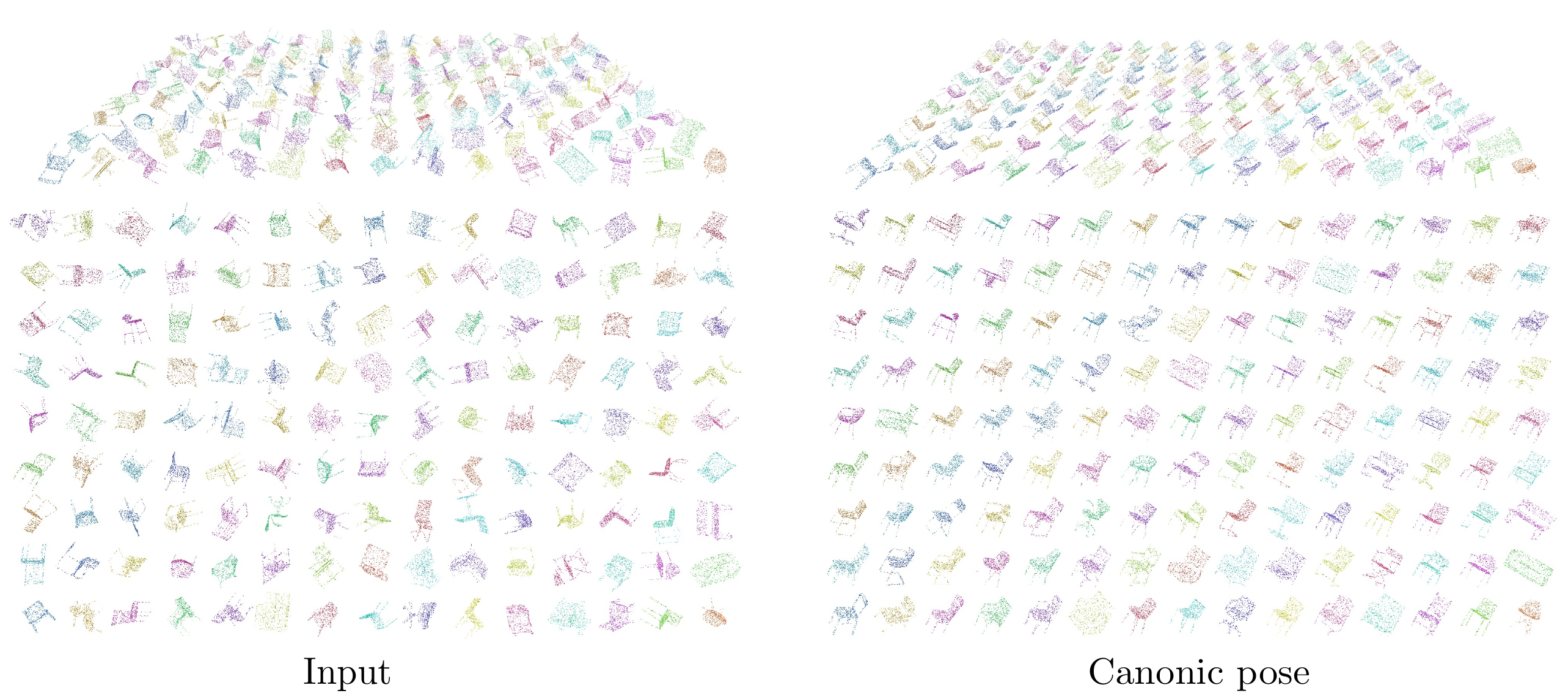}
	\caption{More results of chairs in canonic representations. The chairs on the left are randomly translated and rotated, as seen from a side view (first row) and top view (second row). Our canonical representation, on the right, exhibit good alignment across different instances both in orientation and position.}
	\label{fig:chairs_can}
\end{figure*}

\begin{figure*}[t]
	\centering
	\includegraphics[width=1.0\linewidth]{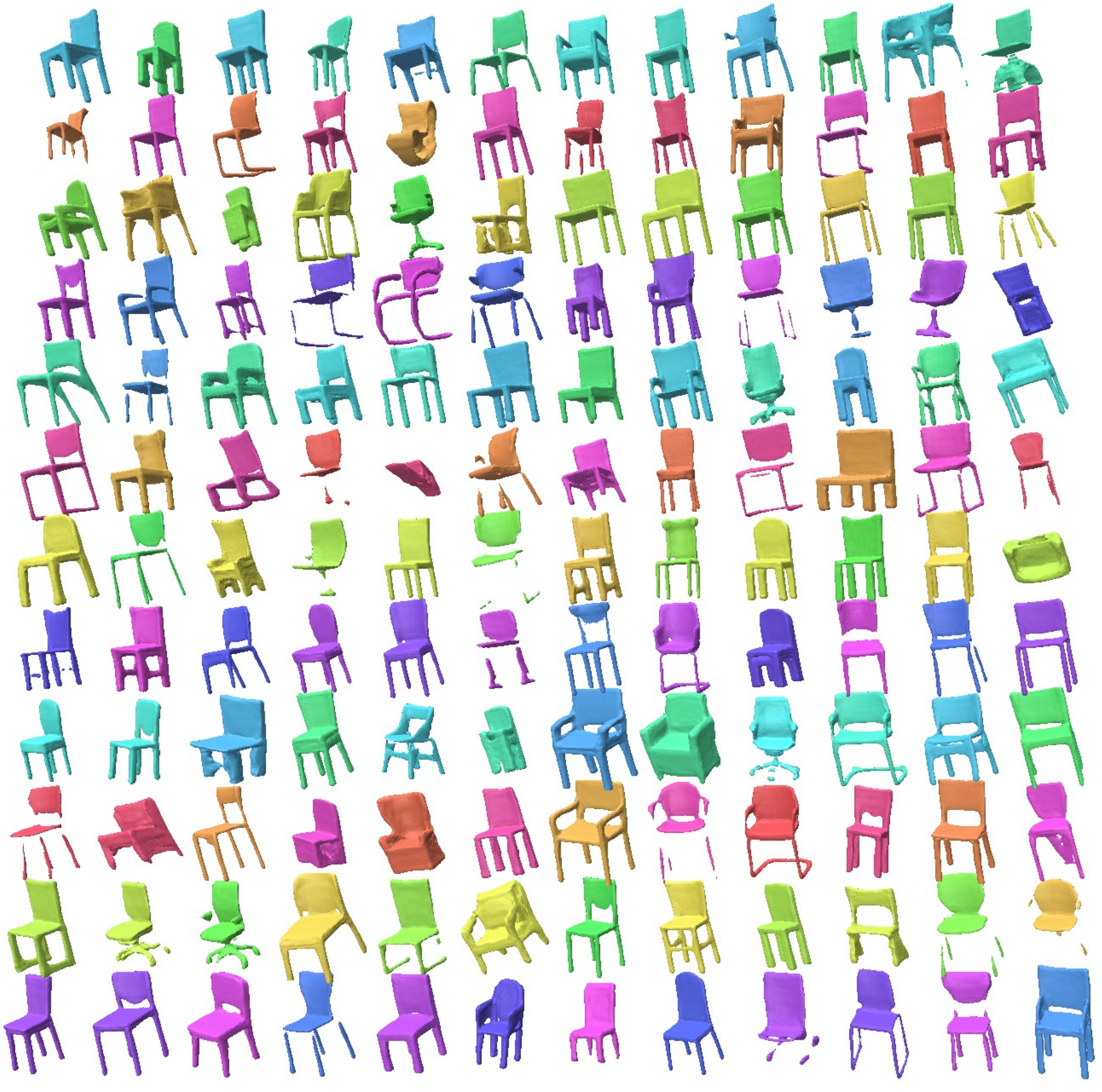}
	\caption{More results of implicit function reconstruction for chairs in canonic representations. Our canonical representation exhibit good alignment across different instances both in orientation and position.}
	\label{fig:chairs_occ_can}
\end{figure*}

\begin{figure*}[t]
	\centering
	\includegraphics[width=1.0\linewidth]{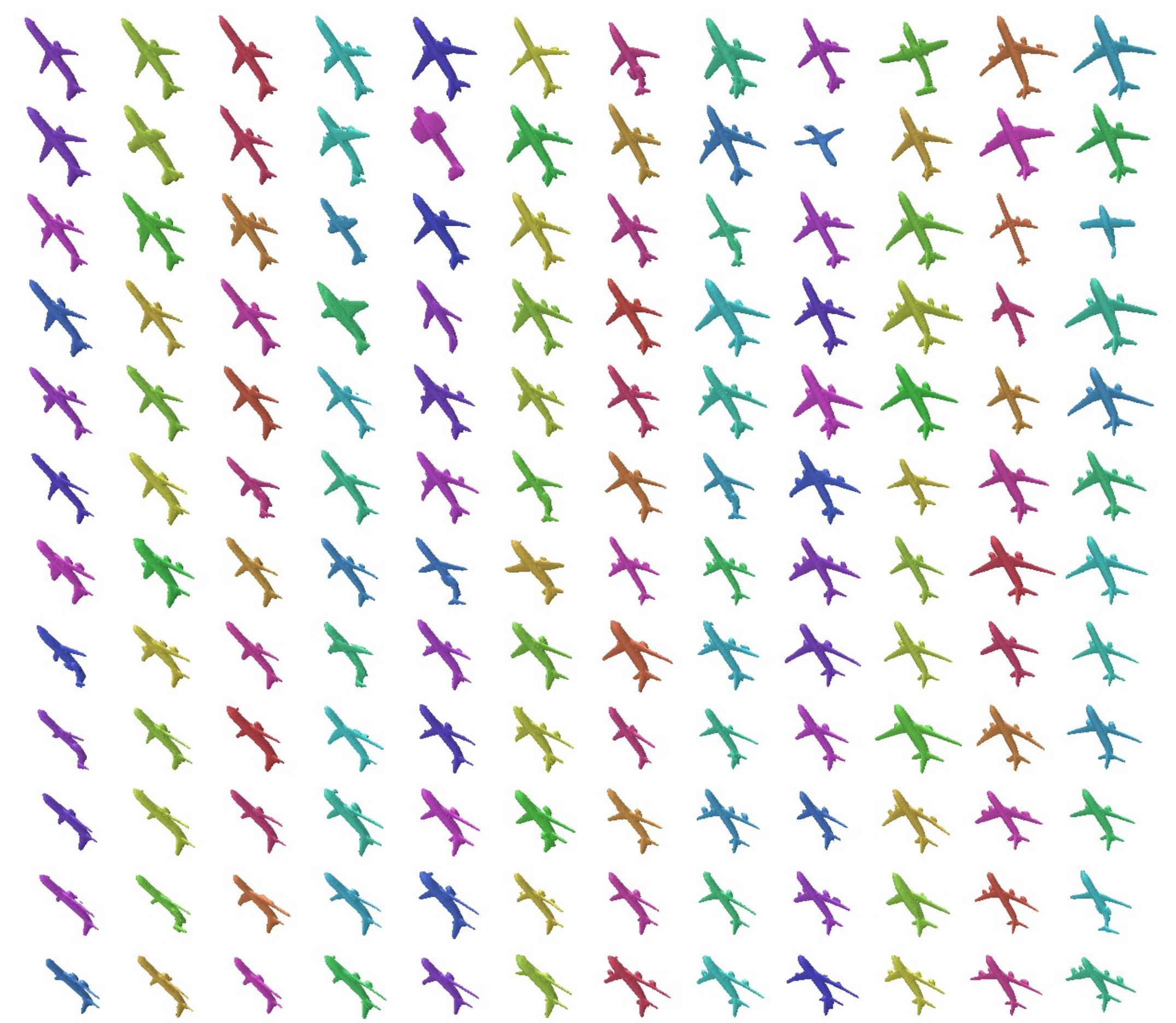}
	\caption{More results of implicit function reconstruction for planes in canonic representations. Our canonical representation exhibit good alignment across different instances both in orientation and position.}
	\label{fig:planes_occ_can}
\end{figure*}

We present more canonical alignment results for point clouds and implicit function reconstruction \figref{fig:planes_can}, \figref{fig:chairs_can}, \figref{fig:chairs_occ_can} and \figref{fig:planes_occ_can}.

\begin{figure*}[t]
	\centering
	\includegraphics[width=1.0\linewidth]{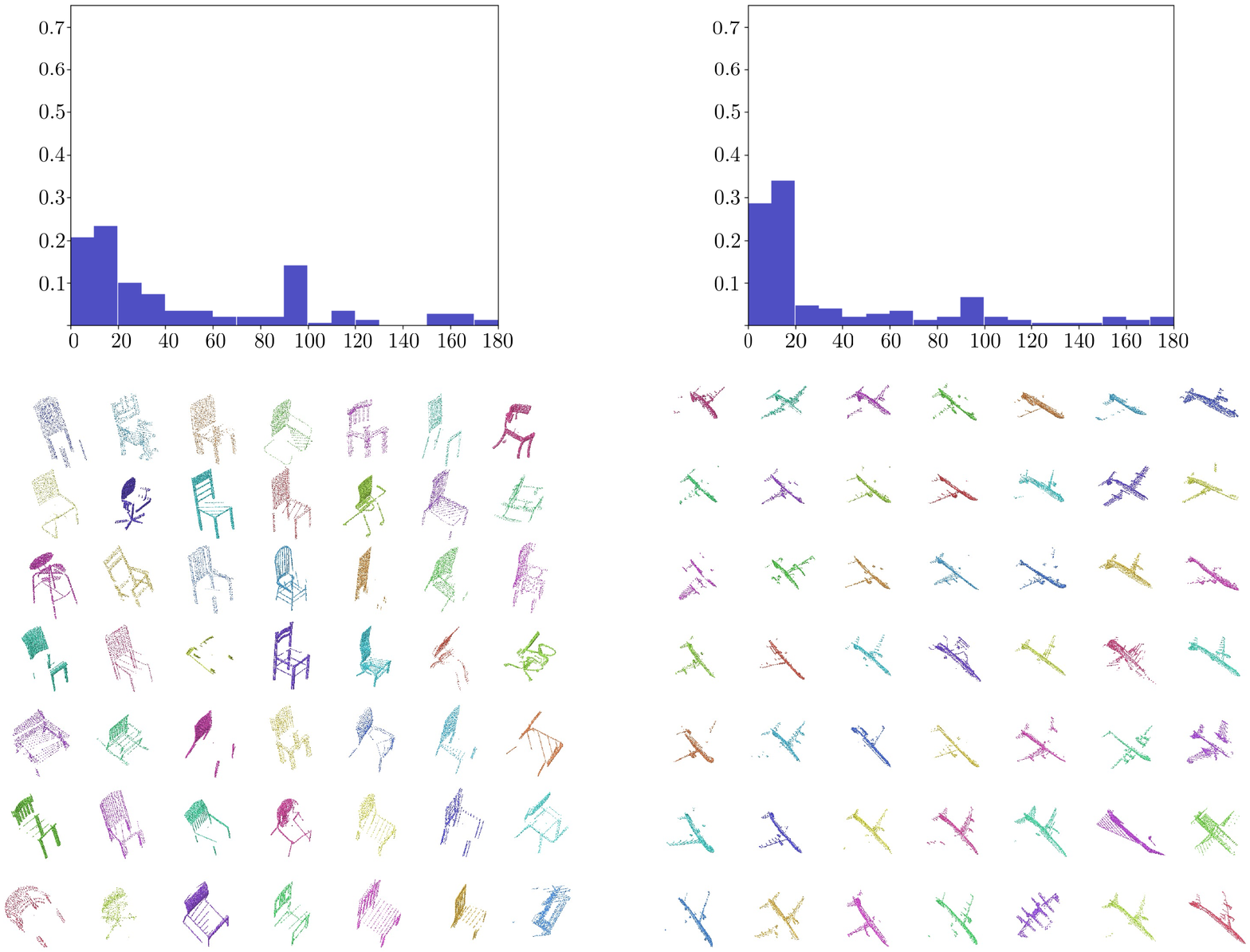}
	\caption{Consistency of the canonical pose for partial shapes of planes and chairs. While our method is not directly optimized to achieve canonic alignment of partial shapes due to occlusions, it has reasonable performances has can be seen in the histogram of the first row and from samples in the second row.}
	\label{fig:partial_can}
\end{figure*}

In addition, we experiment with partial dataset for shape completion derived from ShapeNet (See Yuan et al. "Pcn: Point completion network"). The partial points clouds are a projection of 2.5D depth maps of the model into 3D point clouds. The dataset contains $8$ such partial point clouds per model.
As can be seen in \figref{fig:partial_can} while learning a consistent canonical pose is difficult for partial shapes, our canonical pose is reasonable and mostly consistent. Although, misalignment is apparent in the consistency histogram, please note that no complete point cloud is present in this setting, and no hyper-parameter tuning was done.

\subsection{Stability}
Please see the attached videos for stability visualization, divided to two sub-folders for chairs and airplanes. In each video, we sample a single point cloud and rotate it with multiple random rotation matrices. We feed the rotated point cloud (see on the left of each video) through Compass, Canonical Capsules and Our method, and show the input point cloud in canonical pose for all methods.  Our method reconstruct a \SE-invariant canonical representation and a \SE-equivariant pose estimation, thus, almost no changes are observable in the canonical representation, while both Canonical Capsules and Compass exhibit instability in the canonical pose estimation.
\end{document}